\documentclass[letterpaper]{article}
\usepackage{amsmath}
\usepackage[dvipsnames]{xcolor}
\usepackage{graphicx}
\usepackage{wrapfig}
\usepackage{caption}
\usepackage{subcaption}
\usepackage{amssymb}
\usepackage{booktabs}
\usepackage[draft]{minted}
\usepackage[preprint]{corl_2022} 

\title{Transformers are Adaptable Task Planners}

\author{Vidhi Jain$^{1,2}$  \And Yixin Lin$^1$  \And Eric Undersander$^1$ \And Yonatan Bisk$^2$  \And Akshara Rai$^1$ \\
$^1$ Meta AI, 
$^2$ Carnegie Mellon University 
}

\newcommand{\fullname}{Transformer Task Planner}
\newcommand{\shortname}{TTP}

\begin{document}
\maketitle 

\begin{abstract}
  Every home is different, and every person likes things done in their particular way.
  Therefore, home robots of the future need to both reason about the sequential nature of day-to-day tasks and generalize to user's preferences. To this end, we propose a \fullname{} (\shortname{}) that learns high-level actions from demonstrations by leveraging object attribute-based representations. TTP can be pre-trained on multiple preferences and shows generalization to \textit{unseen} preferences using a single demonstration as a prompt in a simulated dishwasher loading task. 
  Further, we demonstrate real-world dish rearrangement using TTP with a Franka Panda robotic arm, prompted using a single human demonstration. \texttt{Code}: \url{https://anonymous.4open.science/r/temporal_task_planner-Paper148/}
\end{abstract}



\section{Introduction}



Consider a robot tasked with loading a dishwasher. Such a robot has to account for task constraints (e.g. only an open dishwasher rack can be loaded), and dynamic environments (e.g. more dishes may arrive once the robot starts loading the dishwasher). Dishwasher loading is a canonical example of personal preferences, where everyone has a different approach which the robot should adapt to. Classical task planning deals with task constraints through symbolic task description, but such descriptions are difficult to design and modify for new preferences in complex tasks. 
Building easily adaptable long-horizon task plans, under constraints and uncertainty, is an open problem in robotics. 

Machine learning (ML) enables learning complex tasks without extensive expert intervention: robotic navigation \cite{habitatsim2real20ral, truong2020learning, kumar2021rma}, in-hand manipulation \cite{kalashnikov2018qt, nagabandi2020deep, wirnshofer2020controlling, qin2020keto, simeonov2020long}, and planning \cite{yang2020plan2vec, pertsch2020accelerating, singh2020parrot, driess2020deep, andreas2017modular}. Within task planning, ML is used to replace user-defined symbolic descriptions \cite{silver2022inventing}, deal with uncertainty \cite{gordon2019should}, and adapt to preferences \cite{kaushik2020fast}. Recent work \cite{kaplan2017beating} has shown Transformer networks \cite{vaswani2017attention} can learn temporally-consistent representations, and exhibit generalization to new scenarios \cite{brown2020language, sanh2021multitask, liu2021p}. 
Our central question is: \textit{Can a Transformer network learn task structure, adapt to user preferences, and achieve complex long-horizon tasks using no symbolic task representations?}

We hypothesize that task structure and preference are implicitly encoded in demonstrations. 
When loading a dishwasher, a user pulls out a rack before loading it, inherently encoding a structural constraint. They may  place mugs on the top rack and plates on the bottom, encoding their preference. Learning user preferences from long-horizon demonstrations requires policies with temporal context. For example, a user might prefer to load the top rack before the bottom tray. 
The policy needs to consider the \textit{sequence of actions} demonstrated, 
rather than individual actions. 
%
Transformers are well-suited to this problem, as they have been shown to learn long-range relationships \cite{chaplot2021differentiable}, although not in temporal robotic tasks. We propose \fullname{} (\shortname{}) - an adaptation of a classic transformer architecture that includes temporal-, pose- and category-specific embeddings to learn object-oriented relationships over space and time. TTP generalizes beyond what was seen in demonstrations -- to variable numbers of objects and dynamic environments.  By pre-training \shortname{} on multiple preferences, we build temporal representations that can be generalized to new preferences. 
\begin{figure}[t]
\centering
    \includegraphics[width=0.24\textwidth,  trim={ 22cm 0cm 15cm 18cm},clip]{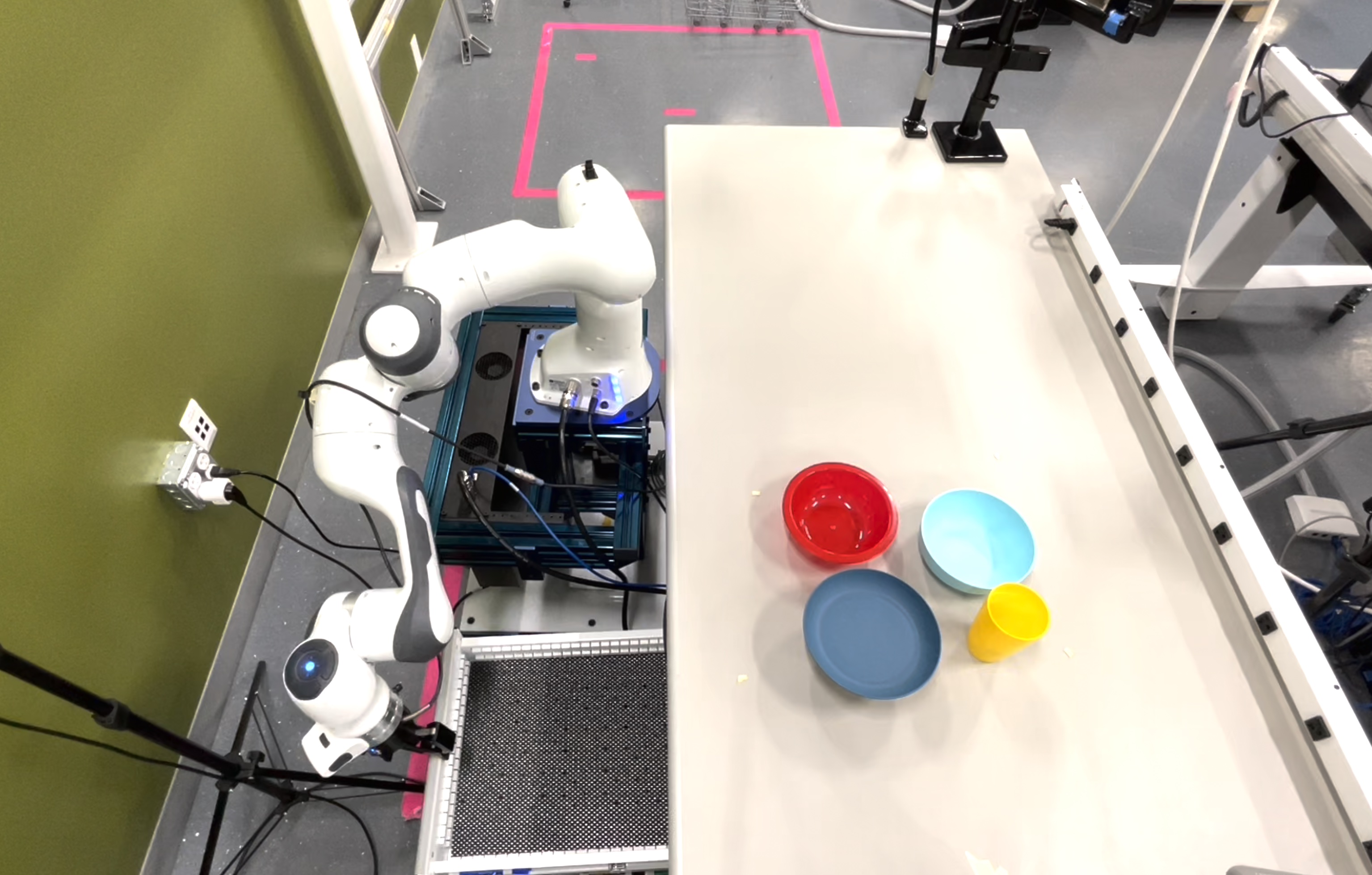} 
    \includegraphics[width=0.24\textwidth,  trim={ 22cm 0cm 15cm 18cm},clip]{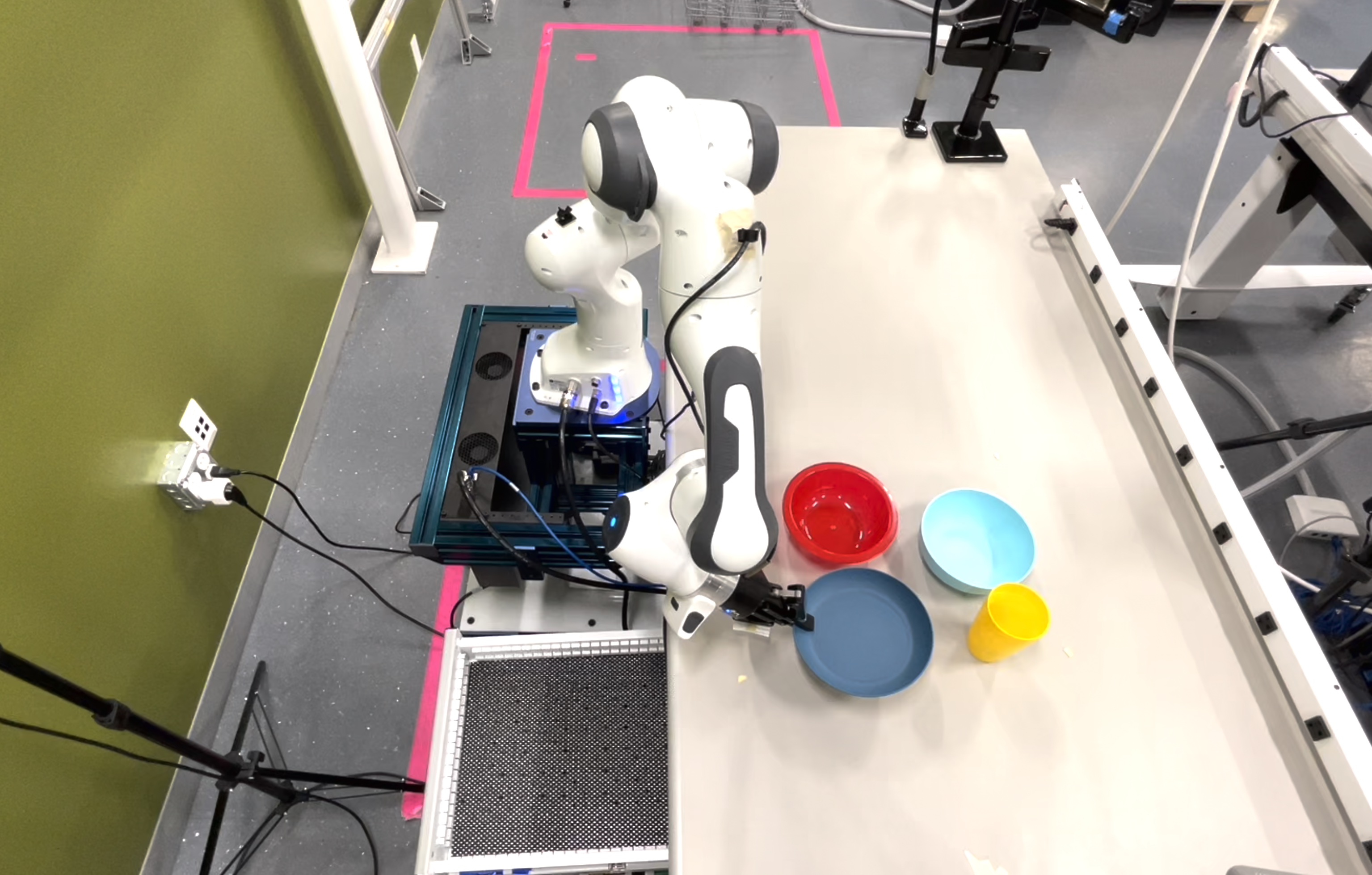}
    \includegraphics[width=0.24\textwidth,  trim={ 22cm 0cm 15cm 18cm},clip]{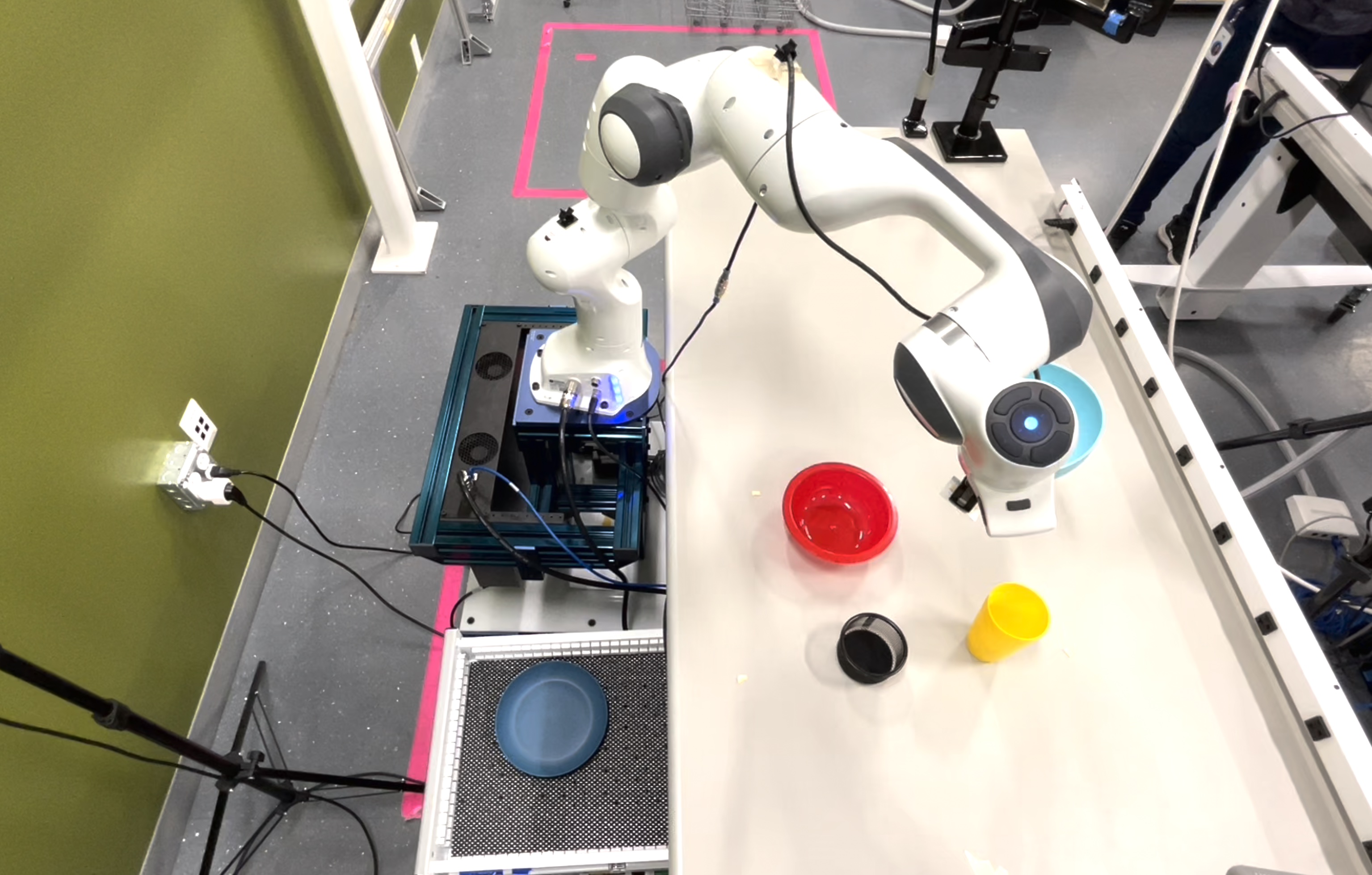} 
    \includegraphics[width=0.24\textwidth,  trim={ 22cm 0cm 15cm 18cm},clip]{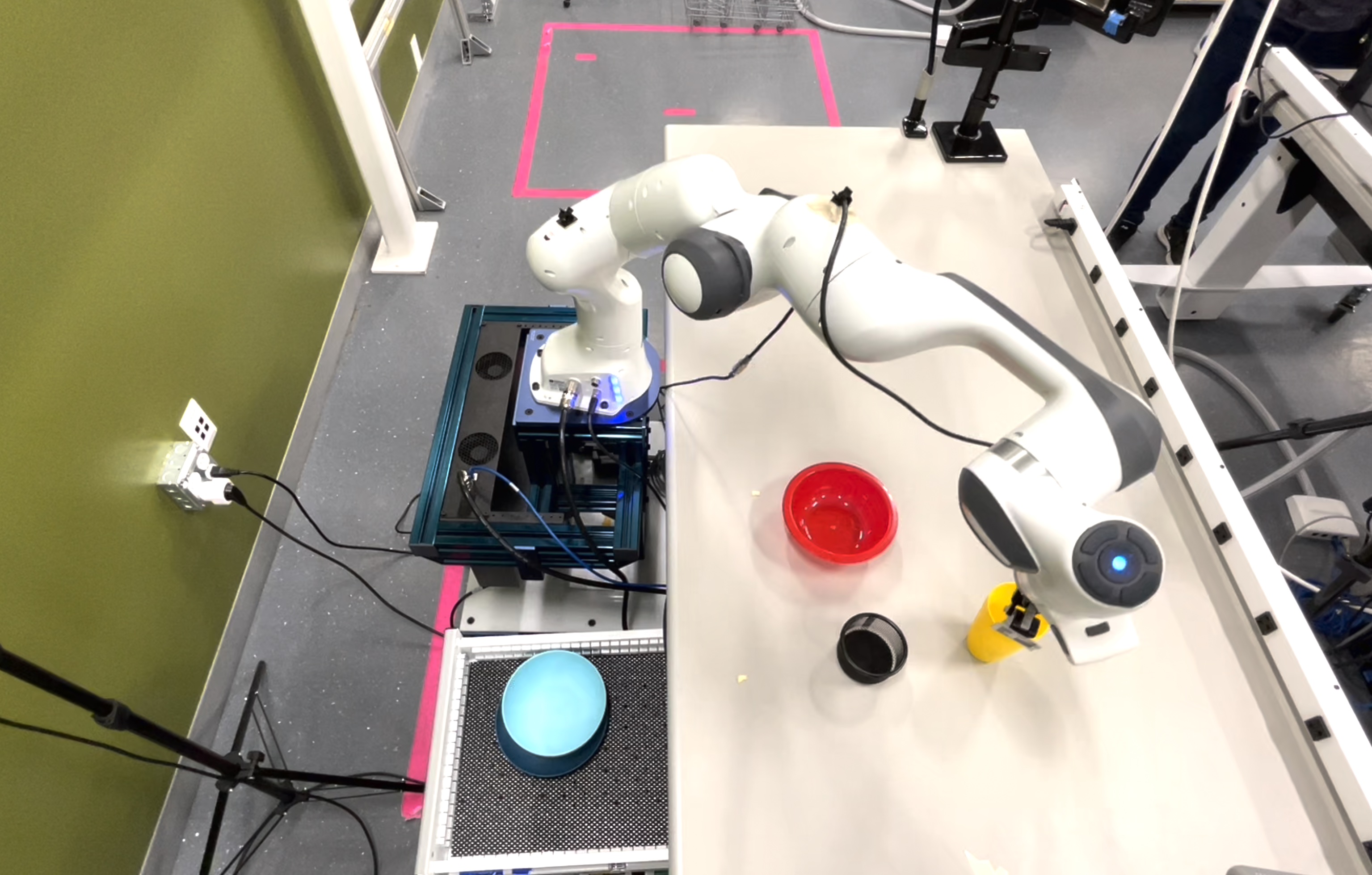} 
    \\
    \includegraphics[width=0.24\textwidth,  trim={ 22cm 0cm 15cm 18cm},clip]{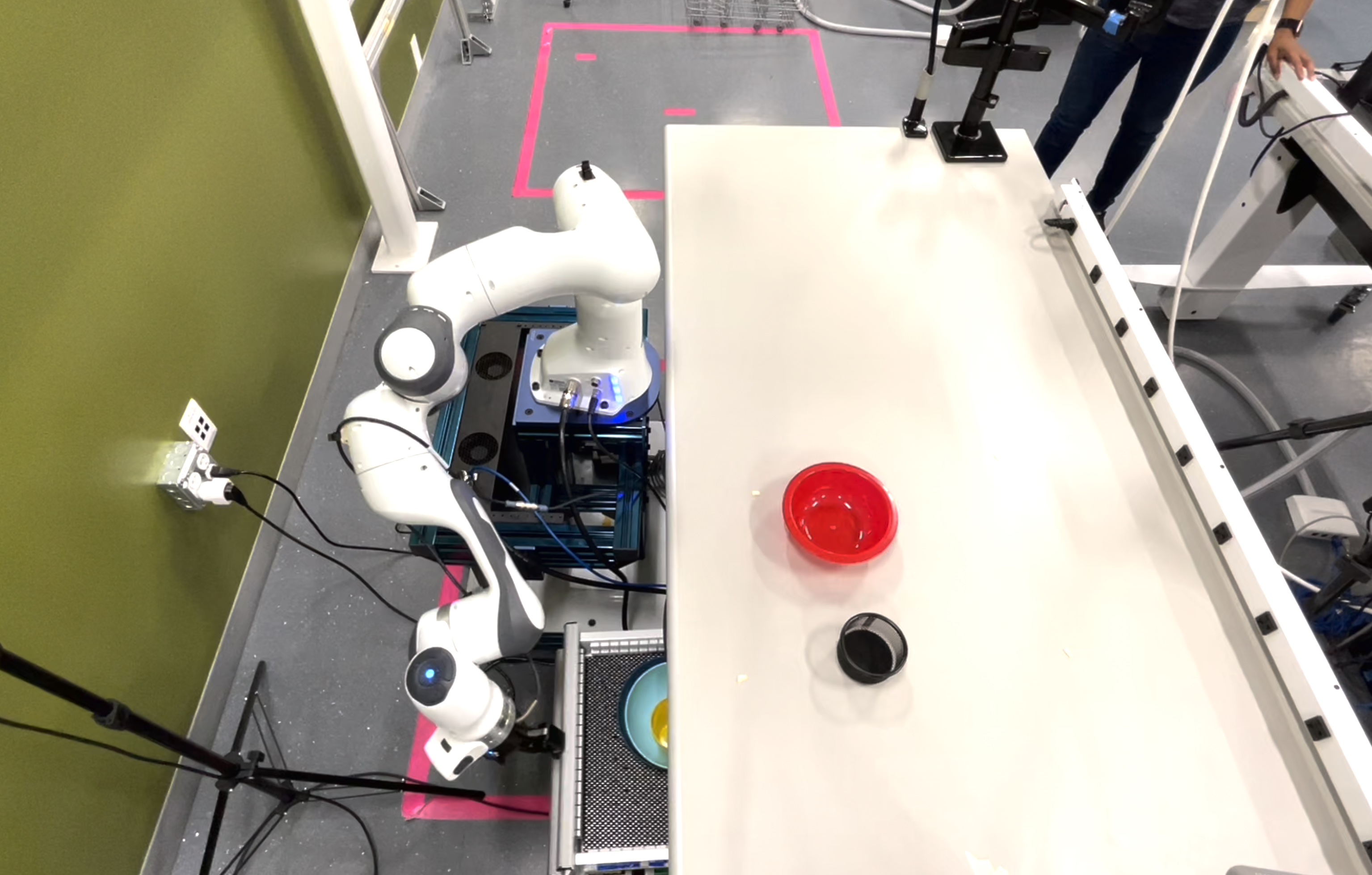}
    \includegraphics[width=0.24\textwidth,  trim={ 22cm 0cm 15cm 18cm},clip]{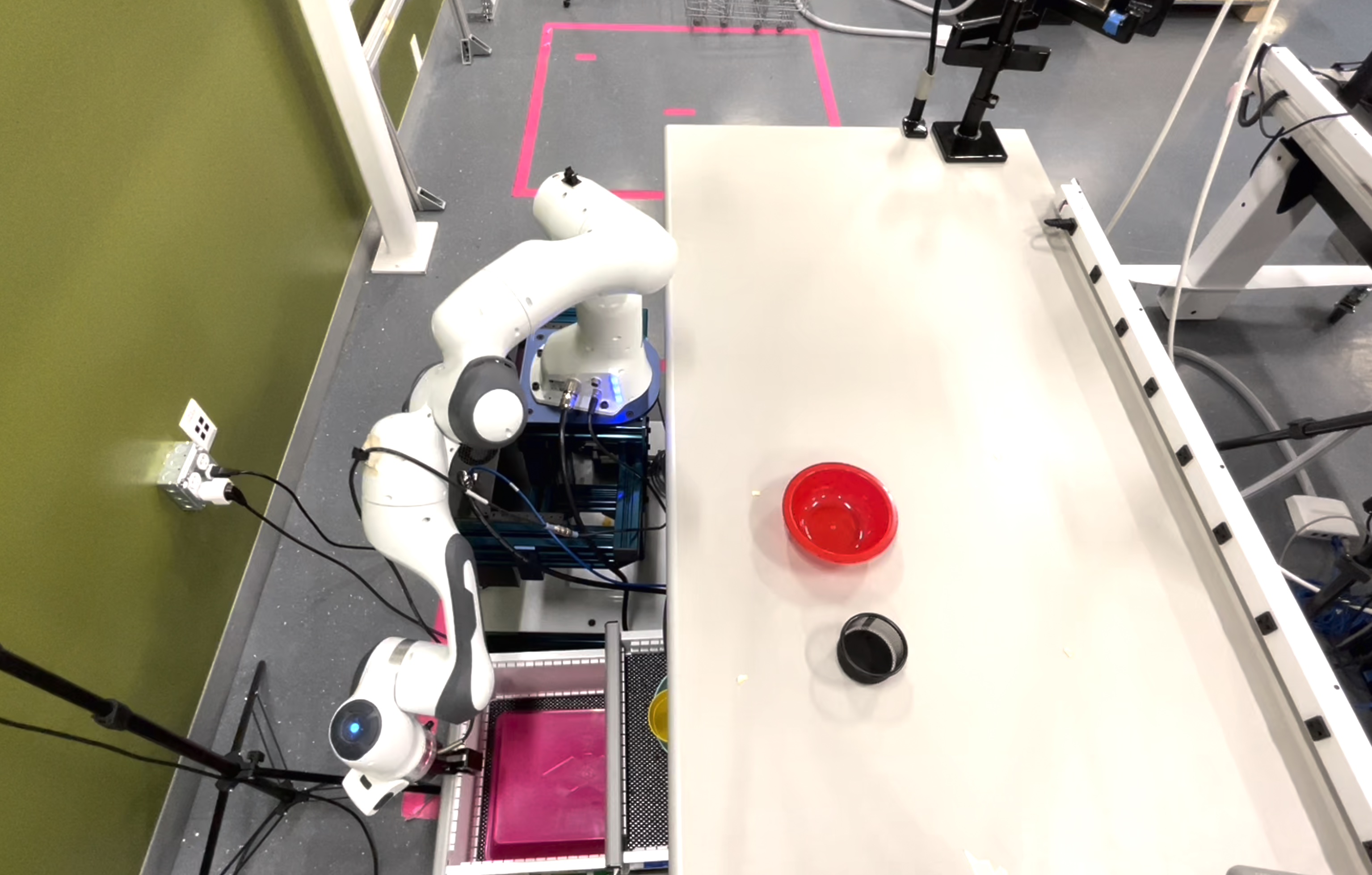} 
    \includegraphics[width=0.24\textwidth,  trim={ 22cm 0cm 15cm 18cm},clip]{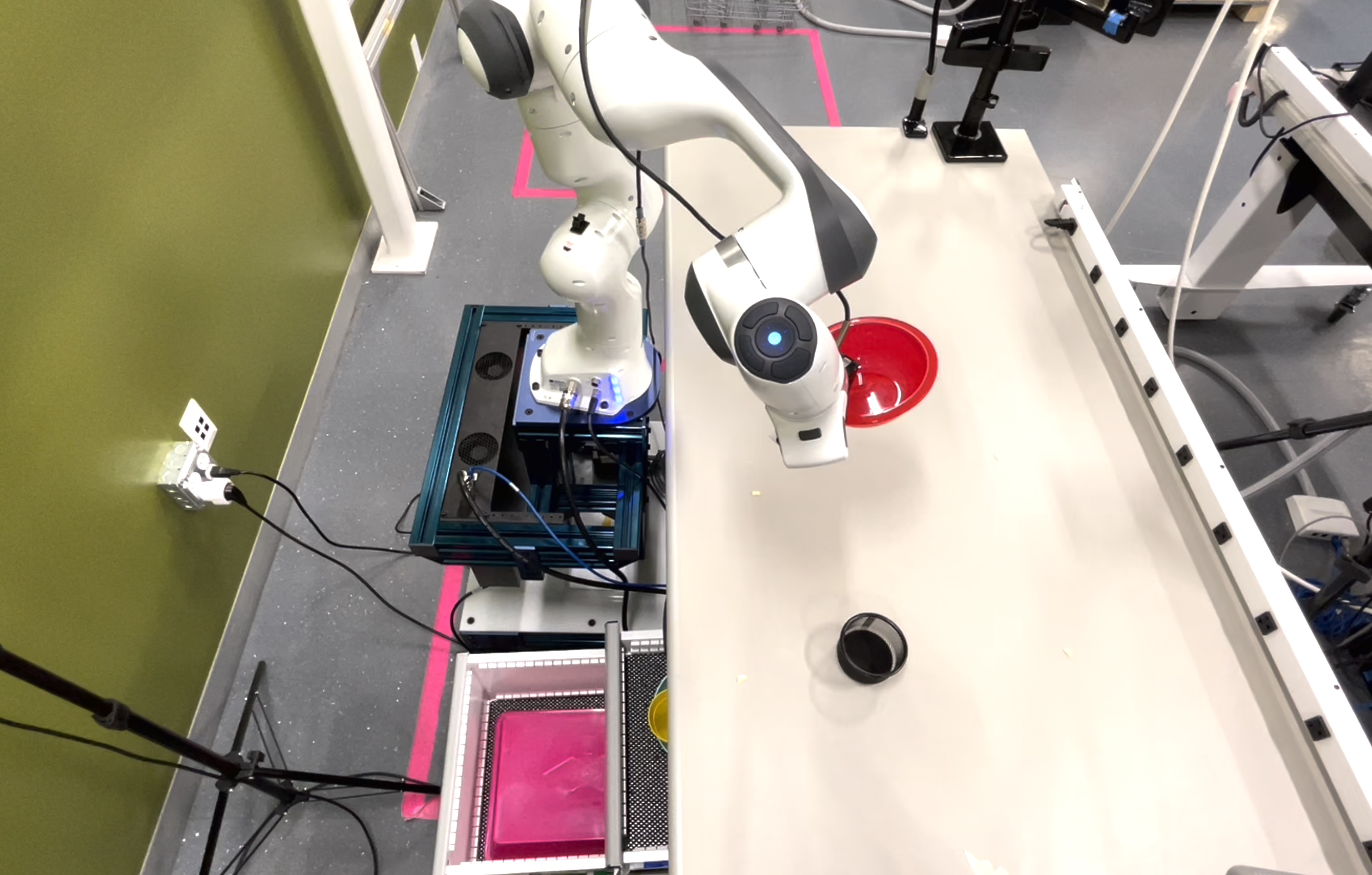}
    \includegraphics[width=0.24\textwidth,  trim={ 22cm 0cm 15cm 18cm},clip]{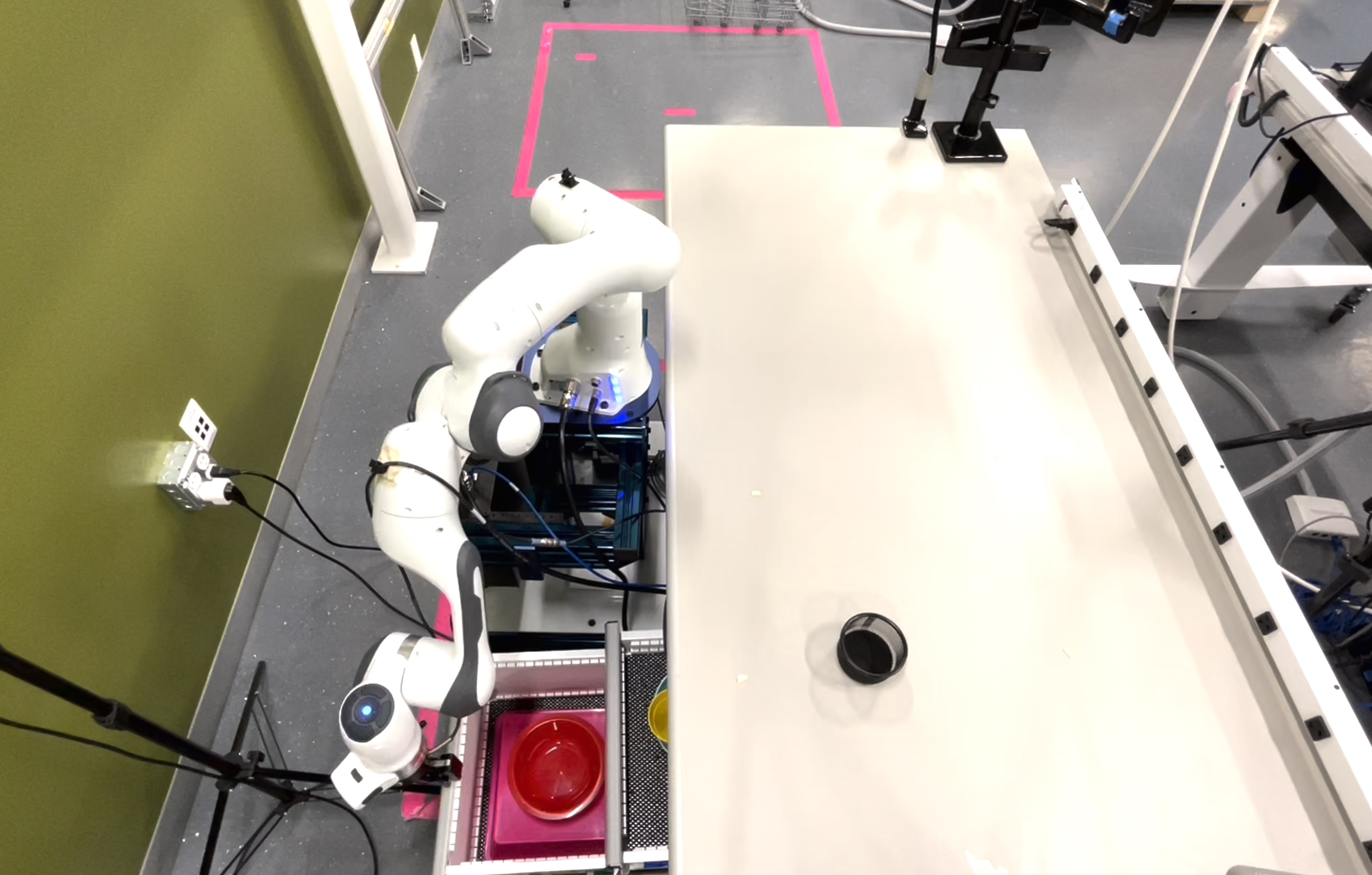} 
    \caption{(Left-to-Right, Top-to-bottom) A Franka Emika Panda arm organizing 4 dishes into 2 drawers, following preference shown in a demonstration. The robot opens a top drawer, places objects in the top drawer, closes it, and does the same for the bottom drawer. 
    The high-level policy that makes decisions about when to open drawers, what objects to pick, etc is learned in simulation and transferred zero-shot to the real-world.
    }
    \label{fig:demo}
\end{figure}
The main contributions of our work are: (1) Introduce transformers as a promising architecture for learning task plans from demonstrations, using object-centric embeddings 
(2) Demonstrate that preference conditioned pre-training generalizes at test time to new, unseen preferences with a single user demonstration. Our experiments use a complex high-dimensional dishwasher loading environment (Fig. \ref{fig:dishwasher_loading}) with several challenges: complex task structure, dynamically appearing objects and human-specific preferences. \shortname{} successfully learns this task from seven preferences with 80 demonstrations each, and generalizes to unseen scenes and preferences and outperforms competitive baselines \cite{lin2022efficient, kapelyukh2022my}. Finally, we transfer \shortname{} to a rearrangement problem in the real-world, where a Franka arm places dishes in two drawers, using a single human demonstration (Fig. \ref{fig:demo}).

\section{Transformer Task Planner (\shortname)}
\label{sec:method}
We introduce \shortname, a Transformer-based policy architecture for learning sequential manipulation tasks. 
We assume low-level `generalized' pick-place primitive actions that apply to both objects like plates, bowls, and also to dishwasher door, racks, etc. \shortname{} learns a high-level policy for pick-place in accordance with the task structure and preference shown in demonstrations. 
The following sections are described with dishwasher-loading as an example, but our setup is applicable to most long-horizon manipulation tasks with `generalized' pick-place.

\begin{figure}
\centering
\begin{subfigure}{0.545\textwidth}
    \centering
    \includegraphics[width=1\textwidth]{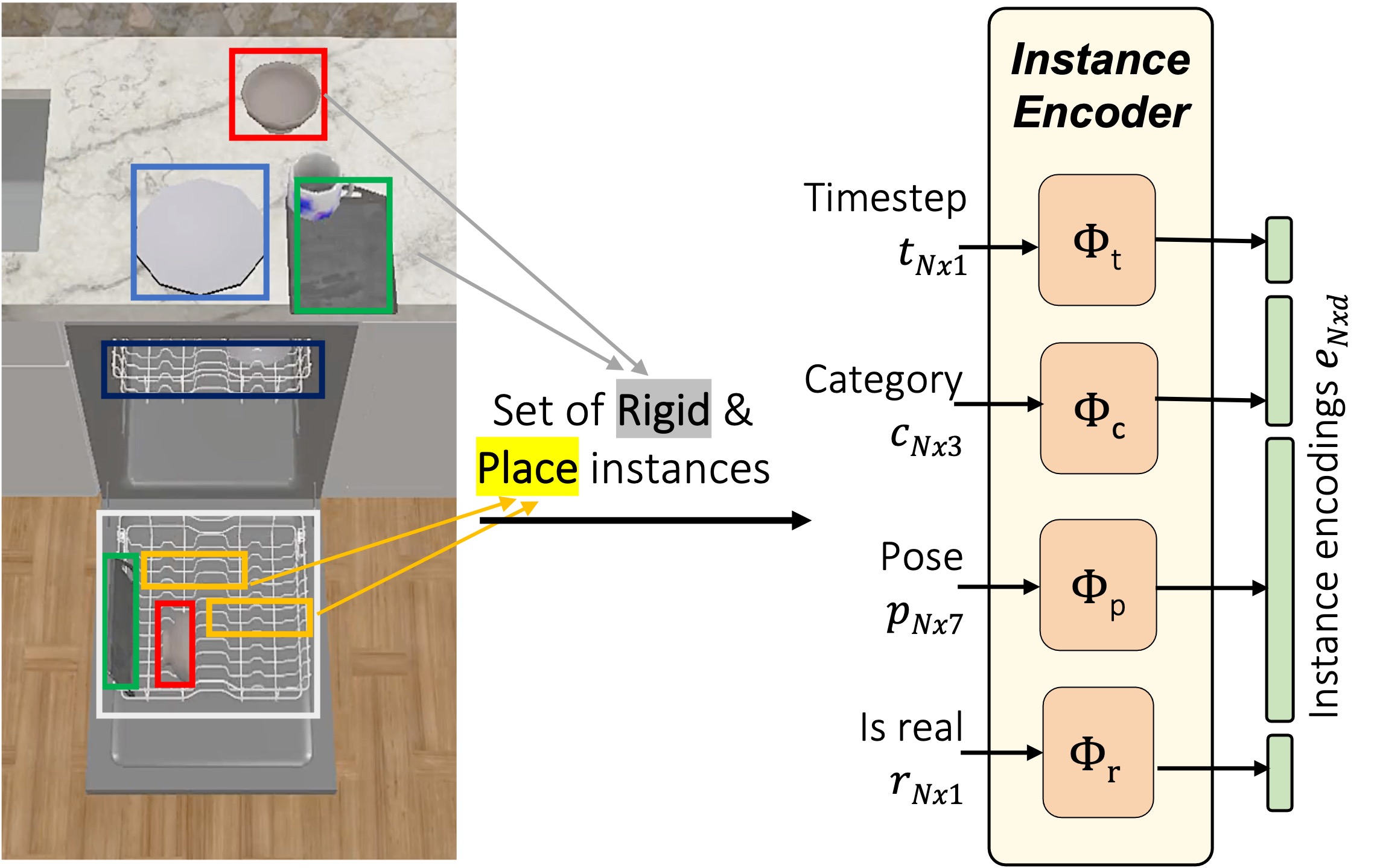}
    \caption{Scene to Instance embeddings}
    \label{fig:scene_to_instance}
\end{subfigure}
\begin{subfigure}{0.445\textwidth}
    \centering
    \includegraphics[width=0.95\textwidth]{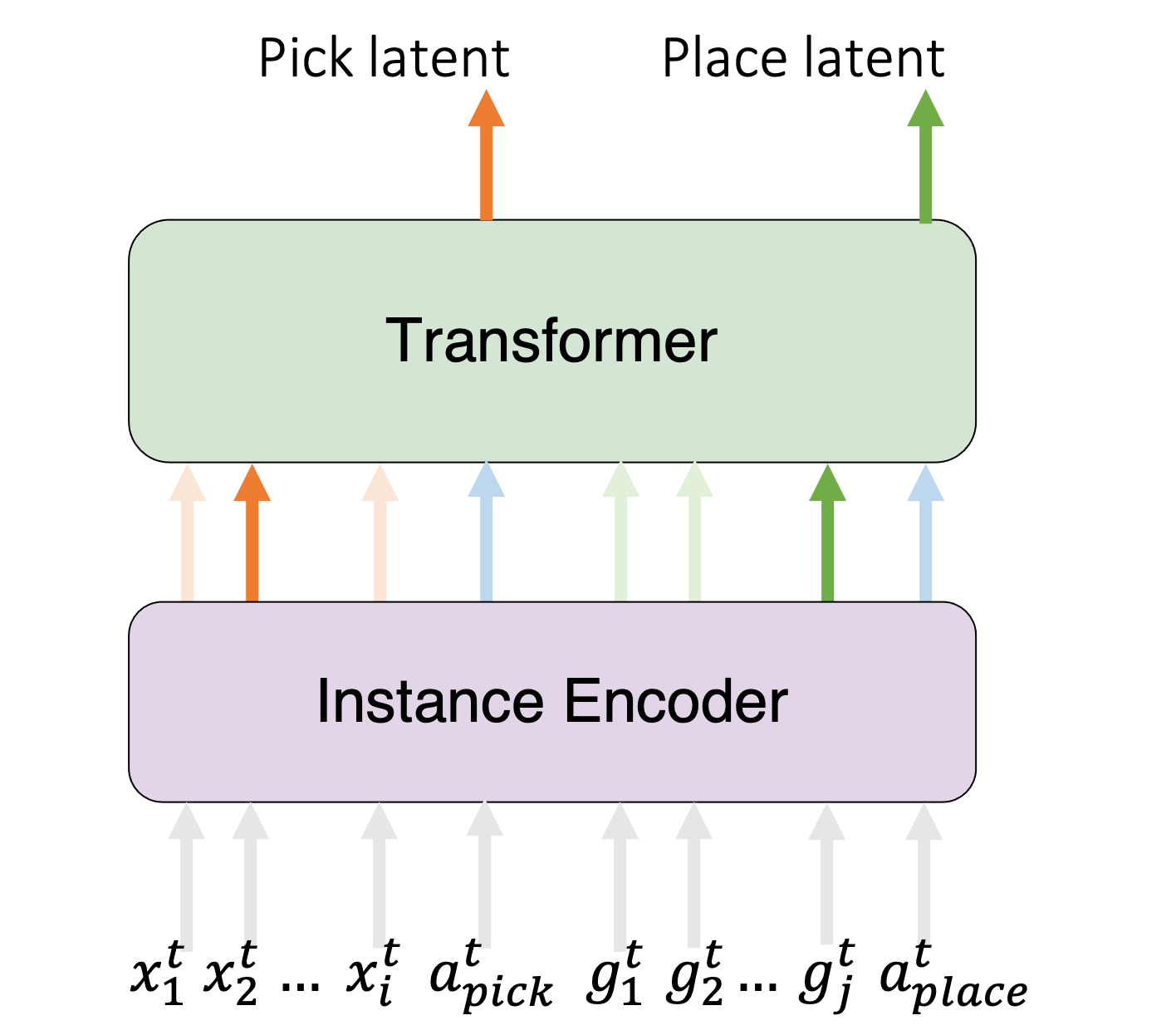}
    \caption{Instances to prediction}
    \label{fig:arch}
\end{subfigure}
\caption{(Left) Architecture overview of how a scene is converted to a set of instances. Each instance is comprised  of attributes, i.e. pose, category, timestep, and whether it is an object or place instance. (Right) Instance attributes (\textcolor{Gray}{gray}) are passed to the encoder, which returns instance embeddings for pickable objects (\textcolor{RedOrange}{red}), placeble locations (\textcolor{OliveGreen}{green}),  and $<$ACT$>$ embeddings (\textcolor{CornflowerBlue}{blue}). The transformer outputs a chosen pick (\textcolor{RedOrange}{red}) and place (\textcolor{OliveGreen}{green})  instance embedding.}
\label{fig:scene2inst-n-inst2pred}
\vspace{-0.6cm}
\end{figure}

\vspace{-0.5em}
\paragraph{State-Action Representations}
We consider a high-level policy that interacts with the environment at discrete time steps. 
At every timestep $t$, we receive observation $\boldsymbol{o}_t$ from the environment which is passed through a perception pipeline to produce a set of rigid-body instances $\{x_i\}^n_{i=1}$, corresponding to the $n$ objects currently visible. We express an \textit{instance} as: $x_i = \{p_i, c_i, t\}$, where $p_i$ is the pose of the object, $c_i$ is its category, and $t$ is the timestep while recording $\boldsymbol{o}_t$ (Fig. \ref{fig:scene_to_instance}). For example, for a bowl at the start of an episode, $p$ is its current location w.r.t a global frame, $c$ is its category (bowl), and $t=0$.
The pick \textit{state} $S^{pick}_t$ is described as the set of instances visible in $\boldsymbol{o}_t:$ 
$S^{pick}_t = \{x_0, x_1,\cdots, x_n\}$. $S^{pick}_t$ is passed to a learned policy $\pi$ to predict a pick action $\boldsymbol{a}_t$. 
We describe the \textit{action} $\boldsymbol{a}_t$ in terms of what to pick and where to place it. Specifically, the pick action chooses one instance from $x_i$ observed in $\boldsymbol{o}_t$: $\pi(S^{pick}) \rightarrow x_{target}, \text{ where } x_{target} \in 
    \{x_i\}^n_{i=1}.$ 

Once a pick object is chosen, a similar 
procedure determines where to place it. We pre-compile a list of discrete placement poses corresponding to viable place locations for each object-category. These poses densely cover the dishwasher racks, and are created by randomly placing objects in the dishwasher and measuring the final pose they land in. 
All possible placement locations for the picked object category, whether free or occupied, are used to create a set of place instances $\{g_j\}_{j=1}^l$. Similar to $x_i$, $g_j = \{p_j, c_j, t\}$, consists of a pose, category and timestep. A boolean value $\mathit{r}$ in the attribute set distinguishes the two instance types.  
The place state is $S^{place} = S^{pick} \cup \{g_j\}_{j=1}^l.$
The \textit{same} policy $\pi$ then chooses where to place the object (Fig.\ref{fig:scene2inst-n-inst2pred}). 
Note that the input for predicting place includes both objects and place instances, since the objects determine whether a place instance is free to place or not. 
$x_{target}$ and $g_{target}$ together make action $\boldsymbol{a}$, sent to a low-level pick-place policy. 
The policy $\pi$ is modeled using a Transformer \cite{vaswani2017attention}.

\vspace{-0.5em}
\paragraph{Instance Encoder} Each object and goal instance $x_i$, $g_j$ is projected into a higher-dimensional vector space (Fig. \ref{fig:scene_to_instance}). Such embeddings improve the performance of learned policies \cite{vaswani2017attention, mildenhall2020nerf}. 
For the pose embedding $\Phi_p$, we use a positional encoding scheme similar to NeRF \cite{mildenhall2020nerf} to encode the 3D positional coordinates and 4D quaternion rotation of an instance. 
For category, we use the dimensions of the 3D bounding box of the object to build a continuous space of object types and process this through an MLP $\Phi_c$. 
For each discrete 1D timestep, we model $\Phi_t$ as a learnable embedding in a lookup table, similar to the positional encodings in BERT \cite{devlin2018bert}. To indicate whether an instance is an object or placement location, we add a 1D boolean value vectorized using a learnable embedding function $\Phi_r$. 
The concatenated $d$-dimensional embedding for an instance at timestep $t$ is represented as $f_e(x_i) = \Phi_t || \Phi_c || \Phi_p || \Phi_r = e_i$.
The encoded state at time $t$ can be represented in terms of instances as: 
$S^{enc}_t = [e_0, e_1, \cdots e_N]$. We drop $()^{enc}$ for brevity.  

\vspace{-0.5em}
\paragraph{Demonstrations}are state-action sequences $\mathcal{C} = \{(S_0, a_0), (S_1, a_1), \cdots, (S_{T-1}, a_{T-1}), (S_{T}) \}$. Here $S_i$ is the set of object and place instances and $a_i$ are the pick-place actions chosen by expert at time $i$. At every time step, we record the state of the objects, the pick instance chosen by the expert, place instances for the corresponding category, and the place instance chosen by the expert. Expert actions are assumed to belong to the set of visible instances in $S_i$. However, different experts can exhibit different preferences over $a_i$. For example, one expert might choose to load bowls first in the top rack, while another expert might load bowls last in the bottom rack. In the training dataset, we assume labels for which preference each demonstration belongs to, based on the expert used for collecting the demonstration. Given $K$ demonstrations per preference $m \in \mathcal{M}$, we have a dataset for preference $m$: $\mathcal{D}_m = \{\mathcal{C}_1, \cdots, \mathcal{C}_K \}$. The complete training dataset consists of demonstrations from all preferences: $\mathcal{D} = \bigcup_{m=1}^M  \mathcal{D}_m$. 
During training, we learn a policy that can reproduce all demonstrations in our dataset. This is challenging, since the actions taken by different experts are different for the same input, and the policy needs to disambiguate the preference. At test time, we generalize the policy to both unseen scenes and unseen preferences. 
\subsection{Prompt-Situation Transformer}

\begin{wrapfigure}{r}{0.44\textwidth}
\centering
    \includegraphics[width=0.44\textwidth]{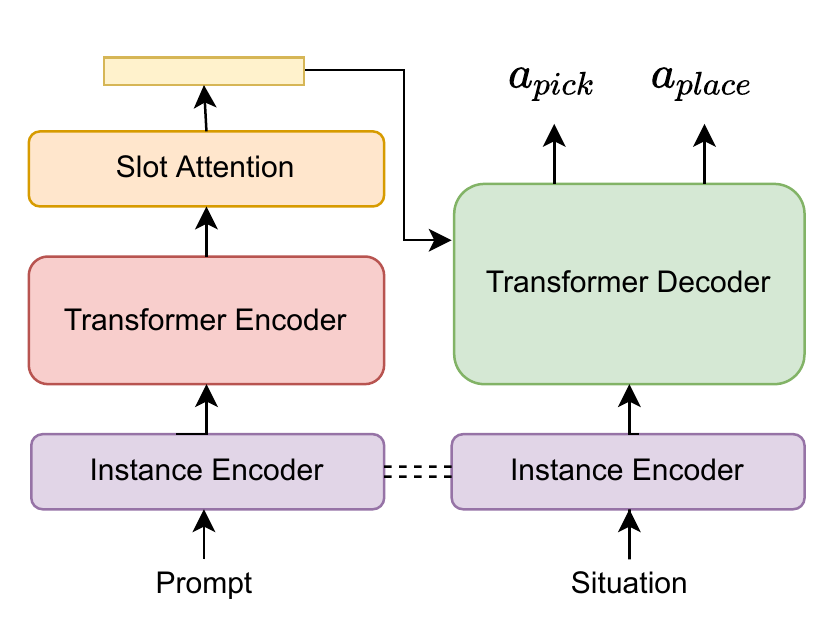}
    \caption{Prompt-Situation Architecture. The left is the prompt encoder which takes as input a prompt demonstration and outputs a learned preference embedding. The right half is the situation decoder, which conditioned on preference embedding from prompt encoder, acts on the current state.}
    \vspace{-0.3cm}
    \label{fig:promptsituation}
    \vspace{-0.2cm}
\end{wrapfigure}

We use Transformers~\cite{vaswani2017attention}, a deep neural network that operates on sequences of data, for learning a high-level pick-place policy $\pi$. 
The input to the encoder is a $d$-dimensional token\footnote{Terminology borrowed from natural language processing where tokens are words; here, they are instances.} per instance $e_i\in[1,..N]$ in the state $S$.
In addition to instances, we introduce special $<$\textsc{ACT}$>$ tokens\footnote{Similar to $<CLS>$ tokens used for sentence classification.}, a zero vector for all attributes, to demarcate the end of one state and start of the next.
These tokens help maintain the temporal structure; all instances between two $<$\textsc{ACT}$>$ tokens in a sequence represent one observed state. 
A trajectory $\tau$, without actions, is: $\tau = \big[S_{t=0}, <$\text{ACT}$>, \cdots, S_{t=T-1}, <$\text{ACT}$>, S_{t=T}\big].$
To learn a common policy for multiple preferences, we propose a prompt-situation architecture (Fig. \ref{fig:promptsituation}). 
The prompt encoder receives one demonstration trajectory as input, and outputs a learned representation of the preference. These output prompt tokens are input to a situation decoder, which also receives the current state as input. The decoder $\pi$ is trained to predict the action chosen by the expert for the situation, given a prompt demonstration.
\label{sec:promt-sit}
The left half is prompt encoder $\psi$ and the right half is the situation decoder or policy $\pi$ acting on given state. 
The prompt encoder $\psi: f_{slot} \circ f_{te} \circ f_{e}$ consists of an instance encoder $f_e$, transformer encoder $f_{te}$, and a slot-attention layer $f_{slot}$ \cite{locatello2020object}. $\psi$ takes the whole demonstration trajectory $\tau_\text{prompt}$ as input and returns a fixed and reduced preference embedding $\gamma = \psi(\tau_\text{prompt})$  of sequence length $H$. 
Slot attention is an information bottleneck, which learns semantically meaningful representations of the prompt.

The situation decoder is a policy $\pi: f_{td} \circ f_e$ that receives as input $N$ instance tokens from the current scene $S$, consisting of objects, as well as, placement instances separated by $<$\textsc{ACT}$>$ tokens (Fig. \ref{fig:arch}). The policy architecture is a transformer decoder \cite{vaswani2017attention} with self-attention layers over the $N$ input tokens. This is followed by a cross-attention layer with preference embedding $\gamma$ ($H$ tokens) from the prompt encoder.
We select the output of the situation decoder at the $<$\textsc{ACT}$>$ token and calculate dot-product similarity with the $N$ input tokens $e_i$. 
The token with the maximum dot-product is chosen as the predicted instance: $x_{pred} = \max_{e_i , i \in \{1, N\}} \big(\hat{x}_{<\textsc{ACT}>} \cdot e_i\big)$. The training target is extracted from the demonstration dataset $\mathcal{D}$, and the policy trained with cross-entropy to maximize the similarity of output latent with the expert's chosen instance embedding. 




\subsection{Multi-Preference Task Learning}
\label{subsec:multi-pref}
We adopt a prompt-situation architecture for multi-preference learning. 
This design (1) enables multi-preference training by disambiguating preferences, (2) learns task-level rules shared between preferences (e.g. dishwasher should be open before placing objects), (3) can generalize to unseen preferences at test time, without fine-tuning. 
Given a `prompt' demonstration of preference $m$, our policy semantically imitates it in a different `situation' (i.e. a  different initialization of the scene). To this end, we learn a representation of the prompt $\gamma^m$ which conditions $\pi$ to imitate the expert.
\begin{align}
    \psi(\tau_\text{prompt}^m)&\rightarrow \gamma^m \\
    \pi(S_\text{situation} | \gamma^m) &
    \rightarrow a^{m}_t = \{x_\text{pred}^m, g_\text{pred}^m\}
\end{align}

%
We train neural networks $\psi$ and $\pi$ together to minimize the total prediction loss over all preferences using the multi-preference training dataset $\mathcal{D}$. The overall objective becomes:
\begin{equation}
    \min_{m \sim M, \tau \sim \mathcal{D}_m, (S,a) \sim\mathcal{D}_m} \mathcal{L}_{CE} (a , \pi(S, \psi(\tau)))
\end{equation}
For every preference $m$ in the dataset, we sample a demonstration from $\mathcal{D}_m$ and use it as a prompt for all state-action pairs $(S,a)$ in $\mathcal{D}_m$. This includes the state-action pairs from $\tau_\text{prompt}$ and creates a combinatorially large training dataset. 
%
At test time, we record one prompt demo from a seen or unseen preference and use it to condition $\pi$ and $\psi$ : $a = \pi(S, \psi(\tau_\text{prompt}))$. All policy weights are kept fixed during testing, and generalization to new preferences is zero-shot using the learned preference representation $\gamma$. Unlike \cite{kapelyukh2022my}, $\gamma$ captures not just the final state, but a temporal representation of the whole demonstration. Building a temporal representation is crucial to encode demonstration preferences like order of loading racks and objects. Even though the final state is the same for two preferences that only differ in which rack is loaded first, our approach is able to distinguish between them using the temporal information in $\tau_\text{prompt}$. To the best of our knowledge, our approach is the first to temporally encode preferences inferred from a demonstration in learned task planners.

\section{Experiments}
\label{sec:experiments}


\begin{figure}[t]
    \centering
    \begin{subfigure}{0.19\textwidth}
    \includegraphics[width=\textwidth]{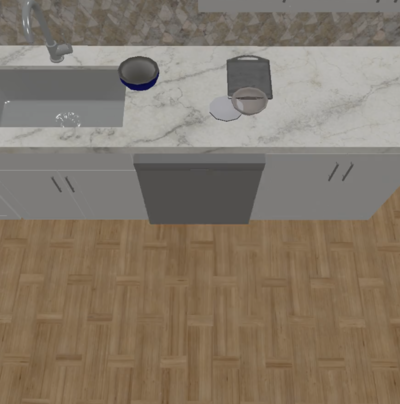}
    \end{subfigure}
    \begin{subfigure}{0.19\textwidth}
    \includegraphics[width=\textwidth]{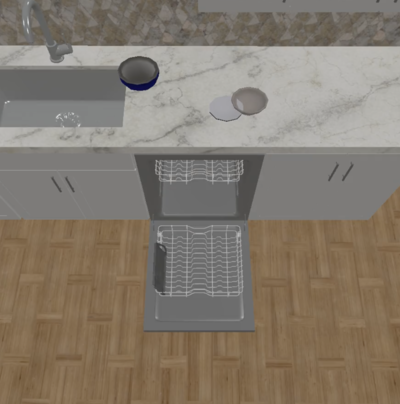}
    \end{subfigure}
    \begin{subfigure}{0.19\textwidth}
    \includegraphics[width=\textwidth]{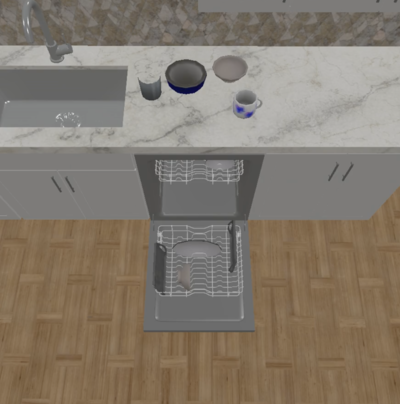}
    \end{subfigure}
    \begin{subfigure}{0.19\textwidth}
    \includegraphics[width=\textwidth]{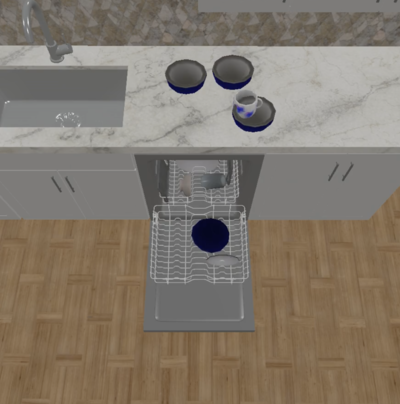}
    \end{subfigure}
    \begin{subfigure}{0.19\textwidth}
    \includegraphics[width=\textwidth]{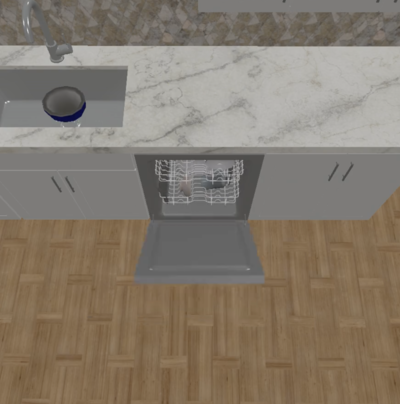}
    \end{subfigure}
    \caption{Dishwasher Loading demonstration in AI Habitat Kitchen Arrange Simulator. Objects dynamically appear on the counter-top (ii-iv), and need to be placed in the dishwasher. If the dishwasher racks are full, they land in sink (v)}
    \label{fig:dishwasher_loading}
    \vspace{-0.5cm}
\end{figure}
We present the ``Replica Synthetic Apartment 0 Kitchen"\footnote{``Replica Synthetic Apartment 0 Kitchen"  was created with the consent of and compensation to artists, and will be shared under a Creative Commons license for non-commercial use with attribution (CC-BY-NC).} (see figure \ref{fig:dishwasher_loading}, appendix and video), an artist-authored interactive recreation of the kitchen of the ``Apartment 0" space from the Replica dataset \cite{replica19arxiv}.
We use selected objects from the ReplicaCAD \cite{szot2021habitat} dataset, including seven types of dishes, and generate dishwasher loading demonstrations using an expert-designed data generation script (see Appendix \ref{appsubsec:dataset}). 
%
Given 7 categories of dishes and two choices in which rack to load first, the hypothesis space of possible preferences is $2\times 7!$. Our dataset consists of 12 preferences (7 train, 5 held-out test) with 100 sessions per preference. In a session, $n \in \{3, ..., 10\}$ instances are loaded in each rack. The training data consists of sessions with 6 or 7 objects allowed per rack. The held-out test set contains 5 unseen preferences and settings for $\{3, 4, 5, 8, 9, 10\}$ objects per rack.
Additionally, to simulate a dynamic environment, we randomly initialize new objects mid-session on the kitchen counter. This simulates situations where the policy does not have full information of every object to be loaded at the start of the session, and has to learn to be reactive to new information. 


We train a 2-head 2-layer Transformer encoder-decoder with 256 input and 512 hidden dimensions, and 50 slots and 3 iterations for Slot Attention (more details in Appendix \ref{appsec:train}).
We test both in- and out-of-distribution performance in simulation. For in-distribution evaluation, 10 sessions are held-out for testing for each training preference. 
For out-of-distribution evaluation, we create sessions with unseen preferences and unseen number of objects. We evaluate trained policies on `rollouts' in the simulation, a more complex setting than accuracy of prediction. Rollouts require repeated decisions in the environment, without any resets. A mistake made early on in a rollout session can be catastrophic, and result in poor performance, even if the prediction accuracy is high. For example, if a policy mistakenly fails to open a dishwasher rack, the rollout performance will be poor, despite good prediction accuracy. 
To measure success, we rollout the policy from an initial state and compare the performance of the policy with an expert demonstration from the same initial state. Note that the policy does not have access to the expert demonstration, and the demonstration is only used for evaluation. Specifically, we measure (1) how well is the final state packed and (2) how much did the policy deviate from the expert demonstration? 

\textbf{Packing efficiency}: We compare the number of objects placed in the dishwasher by the policy to that in the expert's demonstration. 
Let $a_i$ be number of objects in top rack and $b_i$ be objects on bottom rack placed by the expert in the $i^{th}$ demonstration.
If the policy adheres to the preference and places $\hat{a}_i$ and $\hat{b}_i$ on the top and bottom respectively, then the 
$
\textit{packing efficiency (PE)} = \sum_{i} 
\Big( \frac{\hat{a}_i}{max(\hat{a}_i, a_i)} + \frac{\hat{b}_i}{max(\hat{b}_i, b_i)} \Big)
$.
Packing efficiency is between 0 to 1, and higher is better. 
Note that if the policy follows the wrong preference, then PE is 0, even if the dishwasher is full. 

 \textbf{Inverse Edit distance}: We also calculate is the \textit{inverse edit distance} between the sequence of actions taken by the expert versus the learned policy. We compute the Levenshtein distance\footnote{ 
\texttt{Levenshtein distance= textdistance(learned\_seq, expert\_seq) / len(expert\_seq) 
}} (LD) 
 between the policy's and expert's sequence of pick and place instances. Inverse edit distance is defined as $ED = 1 - LD$; higher is better. This measures the temporal deviation from the expert, instead of just the final state. If the expert preference was to load the top rack first and the learned policy loads the bottom first, $PE$ would be perfect, but inverse edit distance would be low.



\subsection{Evaluation on simulated dishwasher loading }
\label{sec:results}

\paragraph{Baselines:} We compare our approach against Graph Neural Network (GNN) based preference learning from \cite{lin2022efficient} and \cite{kapelyukh2022my}. Neither of these works are directly suitable for our task, so we combine them to create a stronger baseline. We use ground-truth preference labels, represented as a categorical distribution and add them to the input features of the GNN, similar to \cite{kapelyukh2022my}. \textbf{For unseen preferences, this is privileged information} that our approach does not have access to. 
We use the output of the GNN to make sequential predictions, as in \cite{lin2022efficient}. Thus, by combining the two works and adding privileged information about ground-truth preference category, we create a GNN baseline which can act according to preference in a dishwasher loading scenario. We train this policy using imitation learning (IL), and reinforcement learning (RL), following \cite{lin2022efficient}. GNN-IL is trained from the same set of demonstrations as TTP using behavior cloning. For GNN-RL, we use Proximal Policy Optimization \cite{schulman2017proximal} from \cite{raffin2021stable}. GNN-RL learns from scratch by directly interacting with the dishwasher environment and obtaining a dense reward. For details, see Appendix \ref{subsec:rewardRL}. 
\begin{figure}
    \centering
    \begin{subfigure}{0.495\textwidth}
    \includegraphics[width=0.9995\textwidth, trim={1cm 0cm 2cm 1cm},clip]{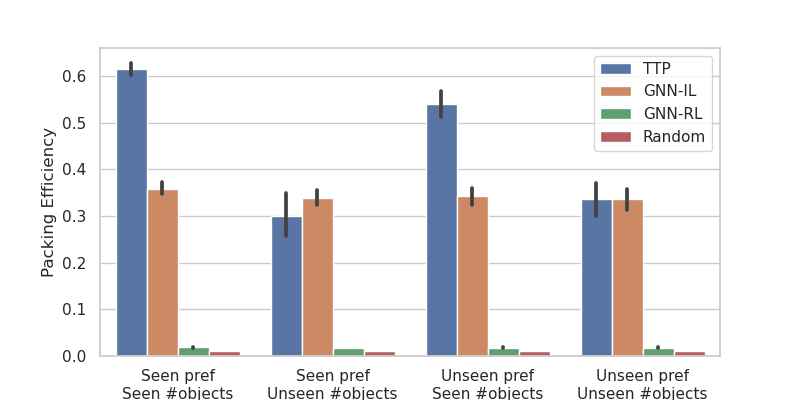}
    \caption{Packing efficiency}
    \end{subfigure}
    \begin{subfigure}{0.495\textwidth}
    \includegraphics[width=0.9995\textwidth, trim={1cm 0cm 2cm 1cm},clip]{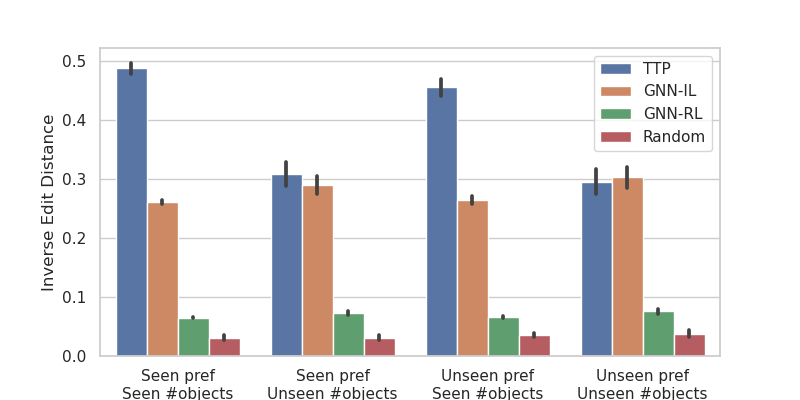}
    \caption{Inverse Edit Distance}
    \end{subfigure}
    \label{fig:results}
    \caption{
    Comparisons of 
    TTP, GNN-IL/RL and a Random policy in simulation across two metrics. TTP shows good performance at the task of dishwasher loading on seen and unseen preferences, and outperforms the GNN and random baselines. However TTP's generalization to unseen \# objects if worse, but still close to GNN-IL, and better than GNN-RL and random.
    }
    \vspace{-0.5cm}
\end{figure}

GNN-IL does not reach the same performance as TTP for in-distribution tasks (PE of $0.34$ for GNN-IL vs $0.62$ for TTP). Note that unseen preferences are also in-distribution for GNN-IL since we provide ground-truth preference labels to the GNN. Hence, there isn't a drop in performance for unseen preferences for GNN, unlike TTP. Despite having no privileged information, TTP outperforms GNN-IL in unseen preferences, and performs comparably on unseen \#objects. Due to the significantly long time horizons per session (more than 30 steps), GNN-RL fails to learn a meaningful policy even after a large budget of 32,000 environment interactions and a dense reward (PE $0.017 \pm 0.002$ using GNN-RL). Lastly, we find that the random policy RP is not able to solve the task at all due to the large state and action space. TTP is able to solve dishwasher loading using unseen preferences well (PE $0.54$). In contrast, classical task planners like \cite{garrett2020integrated} need to be adapted per new preference. This experiment shows that Transformers make adaptable task planners, using our proposed prompt-situation architecture. However, TTP's performance on unseen \#objects deteriorates ($0.62$ for seen versus $0.34$ on unseen \#objects), and we look more closely at that next.

\textbf{Generalization to unseen \# objects}: Fig. \ref{fig:numobj} examines PE on out-of-distribution sessions with lesser i.e. 3-5 or more i.e. 8-10 objects per rack. The training set consists of demonstrations with 6 or 7 objects per rack. The policy performs well on 5-7 objects, but poorer as we go further away from the training distribution. Poorer PE is expected for larger number of objects, as the action space of the policy increases, and the policy is more likely to pick the wrong object type for a given preference. Poorer performance on 3-4 objects is caused by the policy closing the dishwasher early, as it has never seen this state during training. Training with richer datasets, and adding more randomization in the form of object masking might improve out-of-distribution performance of TTP. 

\subsection{Real-world dish-rearrangement experiments}
\label{sec:real-world}
\begin{figure}[t]
    \centering
    \includegraphics[width=\textwidth]{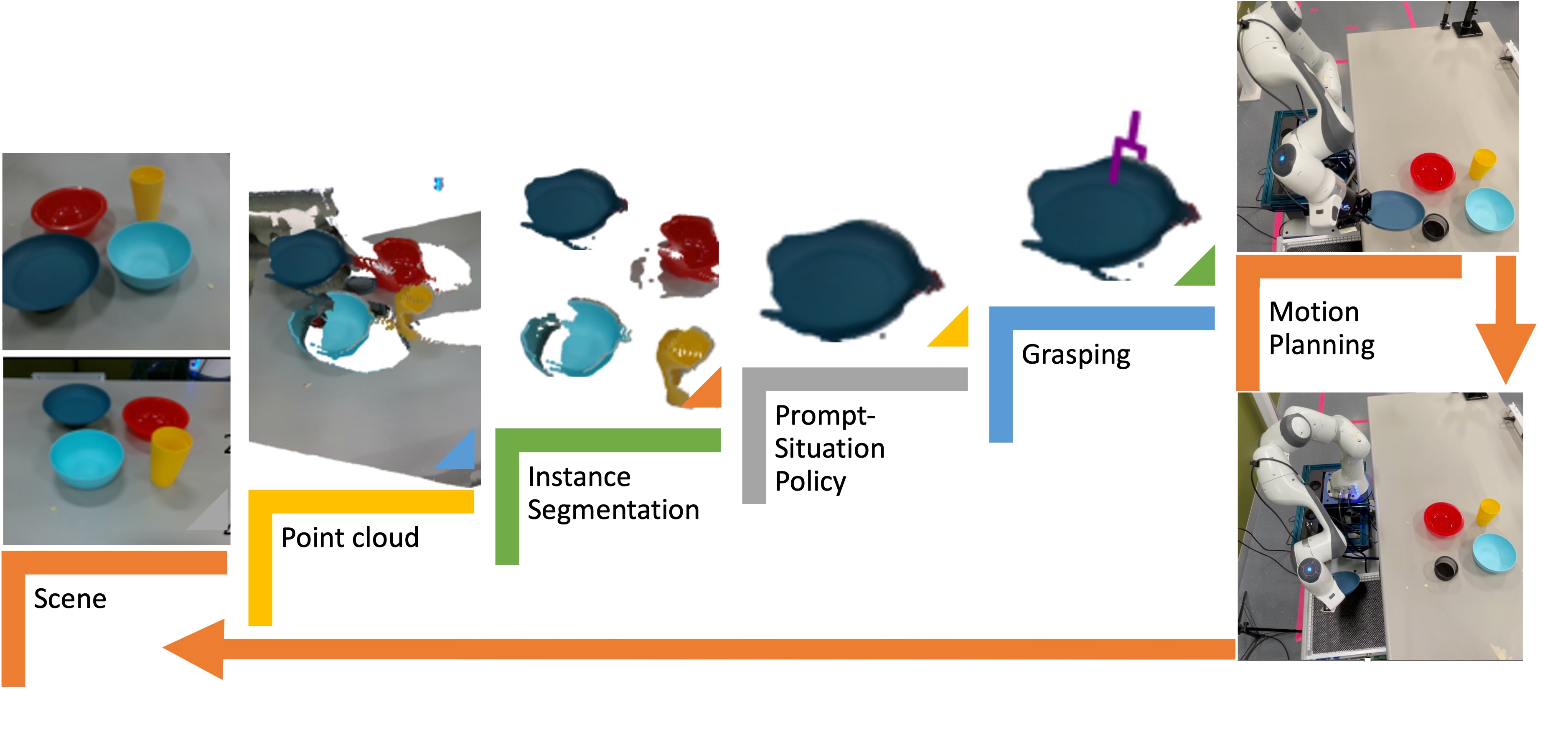}
    \caption{Pipeline for Real Hardware Experiments}
    \label{fig:real_hw_pipeline}
    \vspace{-1em}
\end{figure}
We zero-shot transfer our policy trained in simulation to robotic hardware, by assuming low-level controllers. We use a Franka Panda equipped with a Robotiq 2F-85 gripper, controlled using the Polymetis control framework \cite{Polymetis2021}.
For perception, we find the extrinsics of three Intel Realsense D435 RGBD cameras \cite{keselman2017intel} using ARTags \cite{fiala2005artag}.
The camera output, extrinsics, and intrinsics are combined using Open3D \cite{zhou2018open3d} and fed into a segmentation pipeline \cite{xiang2020learning} to generate object categories. For low-level pick, we use a grasp candidate generator \cite{fang2020graspnet} applied to the object point cloud, and use it to grasp the target object. Place is approximated as a `drop' action in a pre-defined location. Our hardware setup mirrors our simulation, with different categories of dishware (bowls, cups, plates) on a table, a ``dishwasher'' (cabinet with two drawers). The objective is to select an object to pick and place it into a drawer (rack) (see Fig. \ref{fig:demo}).

We use a policy trained in simulation and apply it to a scene with four objects (2 bowls, 1 cup, 1 plate) through the hardware pipeline described above. We start by collecting a prompt human demonstration (more details in Appendix \ref{appsec:hardware_setup})., 
The learned policy, conditioned a prompt demonstration, is applied to two variations of the same scene, and the predicted actions executed.
The policy was successful once with 100\% success rate, and once with 75\%, shown in Figure \ref{fig:demo}, bottom. The failure case was caused due to a perception error -- a bowl was classified as a plate. This demonstrates that such TTP can be trained in simulation and applied directly to hardware. The policy is robust to minor hardware errors, such as if a bowl grasp fails, it just repeats the grasping action (see video and Appendix \ref{appsec:hardware_setup}). However, it relies on accurate perception of the state. In the future, we would like to further evaluate our approach on more diverse real-world settings and measure its sensitivity to the different hardware components,
informing future choices for learning robust policies.



\subsection{Ablations}
\label{sec:ablations}
\begin{figure}[th]
    \begin{subfigure}[b]{0.34\textwidth}
        \includegraphics[width=0.995\textwidth, trim={0.1cm 0cm 1.2cm 1.2cm},clip]{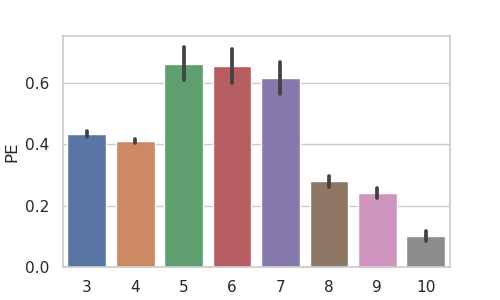}
        \caption{
        Performance decays for OOD 
        number of
        objects, i.e 3-5 \& 8-10}
        \label{fig:numobj}
    \end{subfigure}\hfill
    \begin{subfigure}[b]{0.32\textwidth}
        \centering
        \includegraphics[width=0.995\textwidth, trim={0.5cm 0cm 1.2cm 1.2cm},clip]{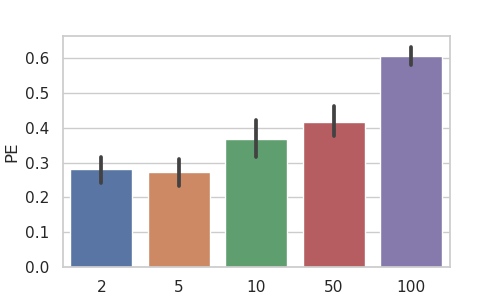}
        \caption{
        Performance improves with the \# of
        unique sessions in training 
        }
        \label{fig:numdemo}
    \end{subfigure} \hfill
    \begin{subfigure}[b]{0.32\textwidth}
        \centering
        \includegraphics[width=0.995\textwidth, trim={0.5cm 0cm 1.2cm 1.2cm},clip]{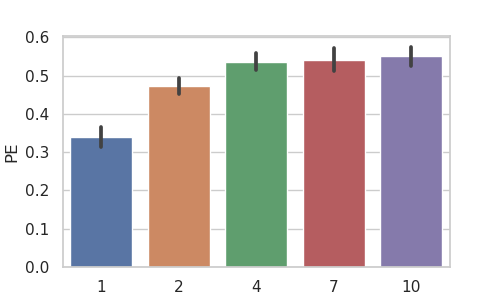}
        \caption{
        Performance improves with the \# of 
        unique preferences in training 
        }
        \label{fig:numpref}
    \end{subfigure}
    \caption{(a) Out-of-distribution generalization to \#objects. (b-c) ablation experiments.}
    \vspace{-0.3cm}
\end{figure}

We study the sensitivity of our approach to training hyperparameters. First, we vary the number of training sessions per preference and study generalization to unseen scenarios of the same preference. Figure \ref{fig:numdemo} shows that the performance of TTP improves as the number of demonstrations increase, indicating that our model is not overfitting to the training set, and might benefit from further training samples. Next, we vary the number of training preferences and evaluate generalization performance to unseen preferences. Figure \ref{fig:numpref} shows that the benefits of adding additional preferences beyond 4 are minor, and similar performance is observed when training from 4-7 preferences. This is an interesting result, since one would assume that more preferences improve generalization to unseen preferences. But for the kinds of preferences considered in our problem, 4-7 preference types are enough for generalization.

\begin{wraptable}[11]{r}{0.3\linewidth}
\vspace{-15pt}
\centering
\caption{Attribute Ablations}
\begin{tabular}{@{}cccl@{}}
\toprule
Pose                  & Cat                  & Time                     & PE \\ \midrule
$\times$                         & $\times$   & $\times$            &           0.0   \\
$\times$                         & $\times$   & \checkmark          &           0.0   \\
$\times$                         & \checkmark & $\times$            &            0.027  \\
$\times$                         & \checkmark & \checkmark          &           0.142   \\
\checkmark & $\times$                         & $\times$                         &   0.411           \\
\checkmark & $\times$                         & \checkmark &       0.419       \\
\checkmark & \checkmark & $\times$                         &  0.517            \\
\checkmark & \checkmark & \checkmark &         0.606 \\ \bottomrule
\end{tabular}
\label{tab:abl_attributes}
\vspace{-0.75cm}
\end{wraptable}

Finally, we analyze which instance attributes are the most important for learning in a object-centric sequential decision making task. We mask different instance attributes to remove sources of information from the instance tokens. 
As seen in Table \ref{tab:abl_attributes}, all components of our instance tokens play significant role ($0.606$ with all, versus the next highest of $0.517$).
The most important attribute is pose of the objects (without the pose, top PE is $0.142$), followed by the category.
The timestep is the least important, but the best PE comes from combining all three.
(more details in Appendix \ref{appsubsec:design_instance}).
\section{Prior Work}
\label{sec:prior_work}




\paragraph{Object-centric representations for sequential manipulation}
Several works build object-centric pick-place representations using off-the-shelf perception methods \cite{wang2019normalized,deng2020self,zeng2017multi,zhu2014single,yoon2003real,jain2021learning}. Once estimated, object states are used by a task and motion planner for sequential manipulation \cite{garrett2020integrated, paxton2019representing}, but objects in the scene are known. 
\cite{florence2019self} combine motor learning with object-centric representation, but transfer of policies is challenging. 
Transporter Nets \cite{zeng2020transporter} use a visual encoder-decoder for table-top manipulation tasks; SORNet \cite{yuan2021sornet} extracts object-centric representations from RGB images and demonstrates generalization in sequential manipulation task. Inspired from these, we learn policies for dishwasher loading, choosing from visible objects to make pick-place decisions.

\vspace{-0.5em}
\paragraph{Transformers for sequence modeling} 
Transformers in NLP \cite{vaswani2017attention} and vision \cite{he2022masked} have focused on self-supervised pretraining due to abundant unsupervised data available. 
Recent works have repurposed transformers for other sequence modeling tasks \cite{Liu2021StructFormerLS,  sun2022plate, jain2020predicting, chen2021decision, janner2021sequence, putterman2022pretraining}. Prompt Decision Transformer \cite{xu2022prompt} considers a single model to encode the prompt and the successive sequence of state. We consider state as variable number of instance 
PlaTe \cite{sun2022plate} proposes planning from videos, while \cite{chen2021decision, janner2021sequence, putterman2022pretraining} model a sequential decision-making task. We consider long-horizon tasks with partially observable state features, and user-specific preferences. 
\vspace{-0.5em}
\paragraph{Preferences and prompt training}
In literature, there are several ways of encoding preferences. \cite{kapelyukh2022my} propose VAE to learn user preferences for spatial arrangement based on just the final state, while our approach models temporal preference from demonstrations.
Preference-based RL learns rewards based on human preferences \cite{wang2022skill, lee2021b, liang2022reward, knox2022models}, but do not generalize to unseen preferences.
On complex long-horizon tasks, modeling human preferences enables faster learning than RL, even with carefully designed rewards \cite{christiano2017deep}.
We show generalization to unseen preferences by using prompts. 
Large language and vision models have shown generalization through prompting \cite{chen2022adaprompt, ramesh2021zero}.
Prompting can also be used to guide a model to quickly switch between multiple task objectives \cite{Raffel2020ExploringTL, Lewis2020BARTDS, Song2019MASSMS}. 
Specifically, language models learn representations that be easy transferred to new tasks in a few-shot setting
\cite{schick2020s, liu2021pre, brown2020language, chen2022adaprompt}. Our approach similarly utilizes prompts for preference generalization in sequential decision-making robotic tasks.
\section{Conclusions and Limitations}
\label{sec:limitations}
We present Tranformer Task Planner (TTP): a high-level, sequential, preference-based policy from a single demonstration using a prompt-situation architecture. We introduced a simulated dishwasher loading dataset with demonstrations that adhere to varying preferences. TTP can solve a complex, long-horizon dishwasher-loading task in simulation and transfer to the real world. 
%

We have demonstrated the TTP's strong performance in the dishwasher setting. This environment is both complex by virtue of its strict sequential nature and yet incomplete as we assume doors and drawers can be easily opened and perception is perfect. In real settings, the policy needs to learn how to recover from its low-level execution mistakes. More complex preferences may depend on differences in visual or textural patterns on objects, for which the instance encoder would require modifications to encode such attributes. An important question to address is how more complex motion plans interact with or hinder the learning objective, especially due to different human and robot affordances.  
Finally, prompts are only presented via demonstration, while language might be a more natural interface for users.





\bibliography{root}
\appendix
\section{Hardware Experiments}
\label{appsec:hardware_setup}




\subsection{Real-world prompt demonstration}

\begin{figure}[h]
    \centering
    \includegraphics[width=0.995\textwidth]{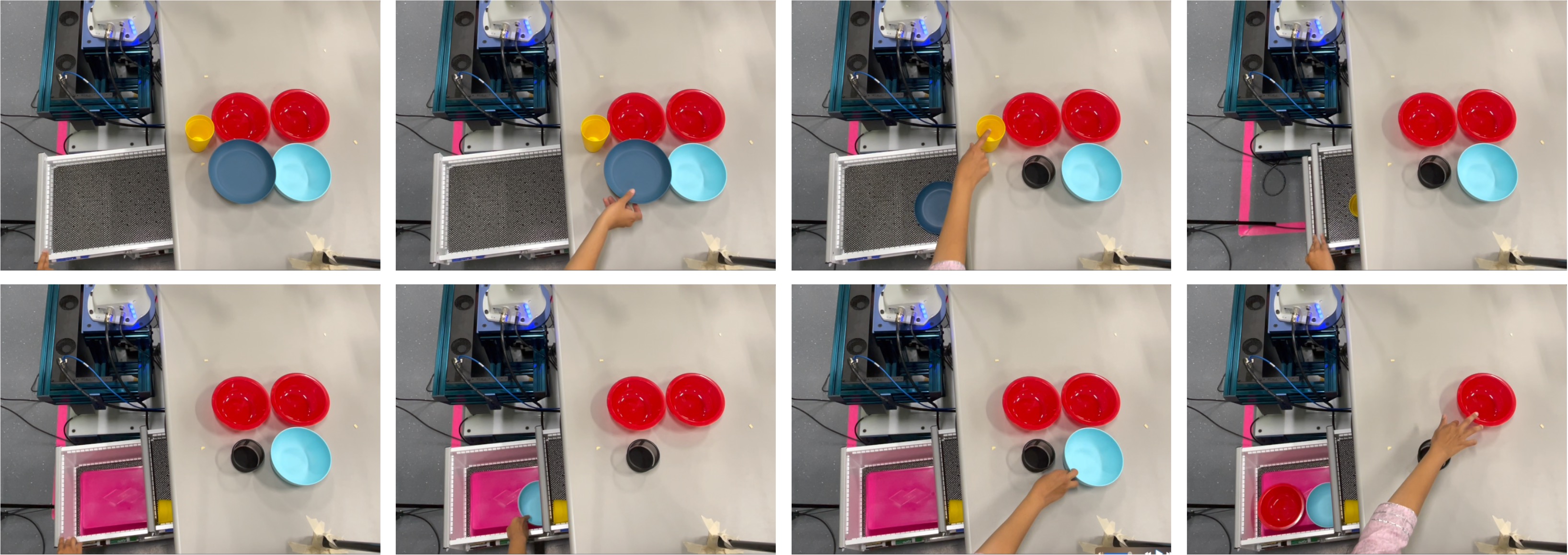}
    \caption{Human demonstration of real-world rearrangement of household dishes.}
    \label{fig:human_prompt_demo}
    \vspace{-3em}
\end{figure}

Here we describe how we collected and processed a visual, human demonstration in the real-world to treat as a prompt for the trained TTP policy (Fig. \ref{fig:human_prompt_demo}). Essentially, we collect demonstration pointcloud sequences and manually segment them into different pick-place segments, followed by extracting object states. At each high-level step, we measure the state using three RealSense RGBD cameras\cite{keselman2017intel}, which are calibrated to the robot frame of reference using ARTags \cite{fiala2005artag}. The camera output, extrinsics, and intrinsics are combined using Open3D \cite{zhou2018open3d} to generate a combined pointcloud. This pointcloud is segmented and clustered to give objects' pose and category using the algorithm from \cite{xiang2020learning} and DBScan. For each object point cloud cluster, we identify the object pose based on the mean of the point cloud.
For category information we use median RGB value of the pointcloud, and map it to apriori known set of objects. In the future this can be replaced by more advanced techniques like MaskRCNN \cite{he2017mask}. Placement poses are approximated as a fixed, known location, as the place action on hardware is a fixed `drop' position and orientation. The per step state of the objects is used to create the input prompt tokens used to condition the policy rollout in the real-world, as described in Section 3.2. 



\subsection{Hardware policy rollout} 

We zero-shot transfer our policy $\pi$ trained in simulation to robotic hardware, by assuming low-level controllers. We use a Franka Panda equipped with a Robotiq 2F-85 gripper, controlled using the Polymetis control framework \cite{Polymetis2021}. Our hardware setup mirrors our simulation, with different categories of dishware (bowls, cups, plates) on a table, a ``dishwasher'' (cabinet with two drawers). The objective is to select an object to pick and place it into a drawer (rack) (see Fig. \ref{fig:human_prompt_demo}).

Once we collect the human prompt demonstration tokens, we can use them to condition the learned policy $\pi$ from simulation. Converting the hardware state to tokens input to $\pi$ follows the same pipeline as the ones used for collecting human demonstrations. At each step, the scene is captured using 3 Realsense cameras, and the combined pointcound is segmented and clustered to get object poses and categories. This information along with the timestep is used to generate instance tokens as described in Section 2 
for all objects visible to the cameras. For visible already placed objects, the place pose is approximated as a fixed location. 
The policy $\pi$, conditioned on the human demo, reasons about the state of the environment, and chooses which object to pick. Next, we use a grasp generator from \cite{fang2020graspnet} that operates on point clouds to generate candidate grasp locations on the chosen object. We filter out grasp locations that are kinematically not reachable by the robot, as well as grasp locations located on points that intersect with other objects in the scene. Next, we select the top 5 most confident grasps, as estimated by the grasp generator, and choose the most top-down grasp. We design an pre-grasp approach pose for the robot which is the same final orientation as the grasp, located higher on the grasping plane. The robot moves to the approach pose following a minimum-jerk trajectory, and then follows a straight line path along the approach axes to grasp the object. Once grasped, the object is moved to the pre-defined place pose and dropped in a drawer. The primitives for opening and closing the drawers are manually designed on hardware.

\begin{figure}[t] 
    \centering
    \includegraphics[width=0.957565\textwidth]{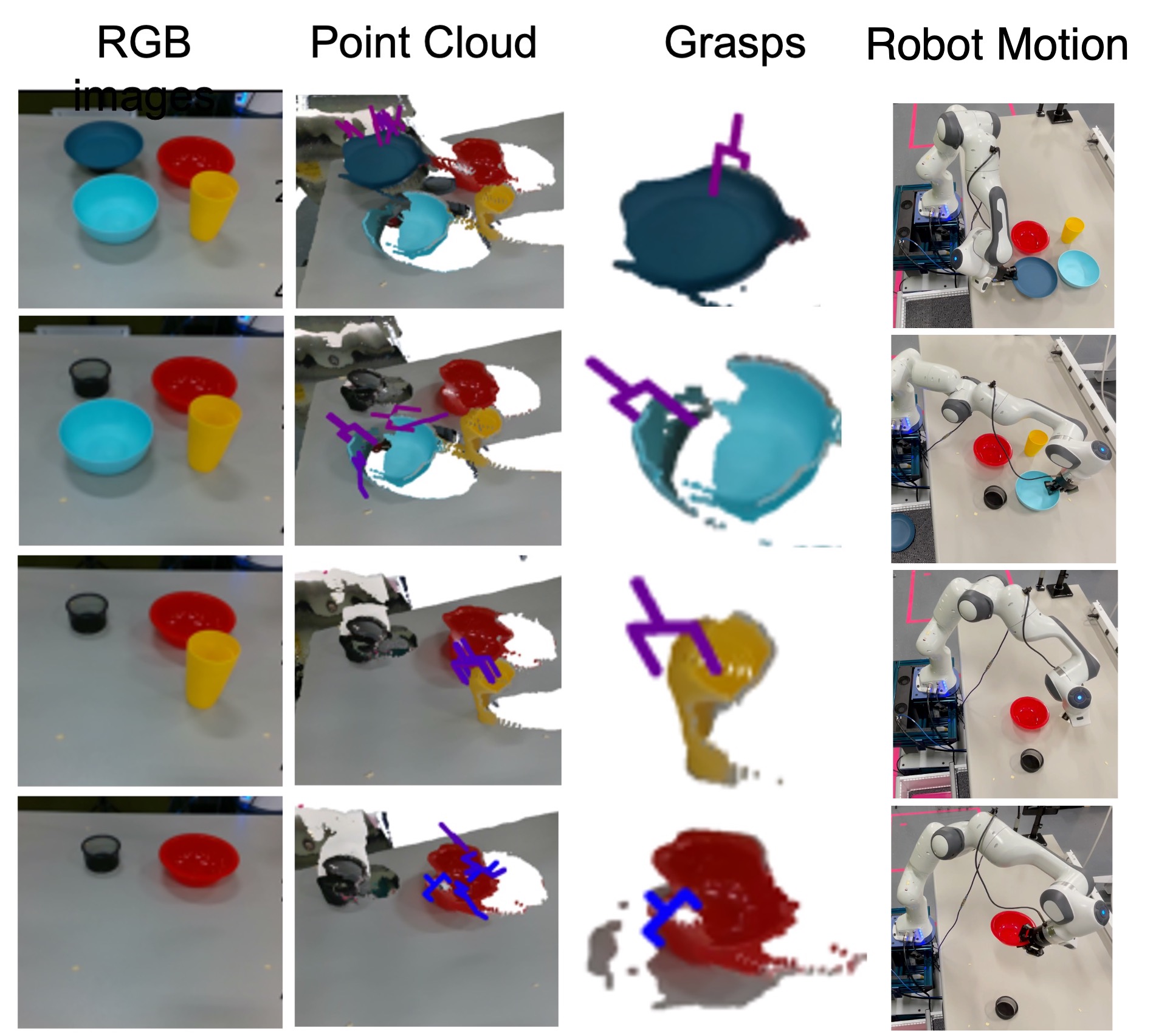}
    \caption{Point cloud and grasps for different objects during policy rollout.} 
    \vspace{-1.5em}
    \label{fig:pcd_n_grasps}
\end{figure}

The learned policy, conditioned on prompt demonstrations, is applied to two variations of the same scene, and the predicted pick actions are executed. Fig.\ref{fig:pcd_n_grasps} shows the captured  image from one of the three cameras, the merged point cloud and the chosen object to pick and selected grasp for the same. 
The policy was successful once with 100\% success rate, and once with 75\%, shown in Fig.\ref{fig:demo}. The failure case was caused due to a perception error -- a bowl was classified as a plate. This demonstrates that our approach (TTP) can be trained in simulation and applied directly to hardware. The policy is robust to minor hardware errors like a failed grasp; it just measures the new state of the environment and chooses the next object to grasp. For example, if the robot fails to grasp a bowl, and slightly shifts the bowl, the cameras measure the new pose of the bowl, which is sent to the policy. However, TTP relies on accurate perception of the state. If an object is incorrectly classified, the policy might choose to pick the wrong object, deviating from the demonstration preference. In the future, we would like to further evaluate our approach on more diverse real-world settings and measure its sensitivity to the different hardware components,
informing future choices for learning robust policies.

\subsection{Transforming hardware to simulation data distribution}

\begin{wrapfigure}{l}{0.35\textwidth}
    \vspace{-1em}
    \includegraphics[width=0.35\textwidth]{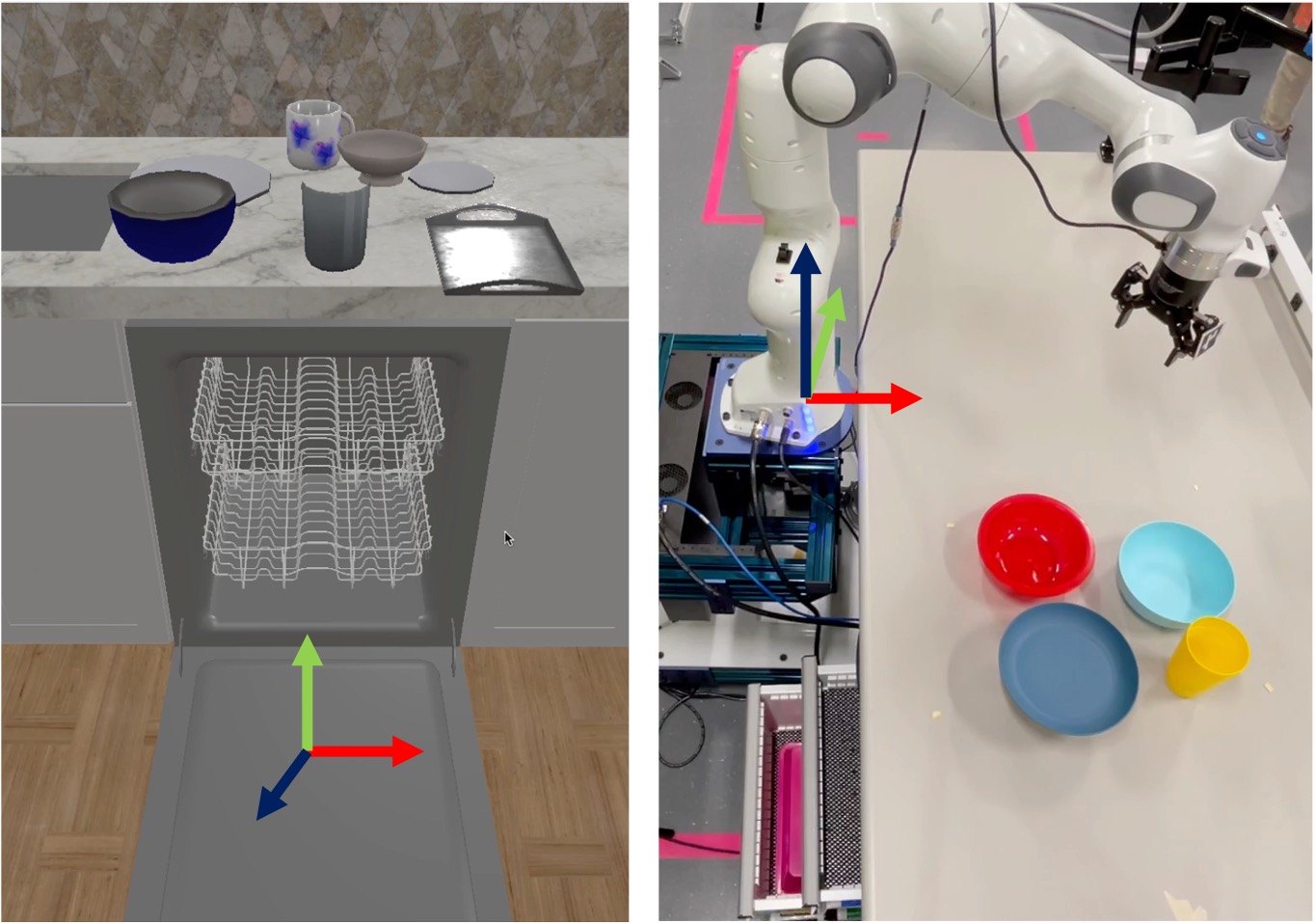}
    \caption{Coordinate Frame of reference in simulation (left) and real world setting (right). Red is x-axis, green is y-axis and blue is z-axis.}
    \label{fig:frame_of_ref}
\end{wrapfigure}

The policy trained in simulation applies zero-shot to real-world scenarios, but it requires a coordinate transform. Fig. \ref{fig:frame_of_ref} shows the coordinate frame of reference in simulation and real world setting.   Since our instance embedding uses the poses of objects, it is dependant on the coordinate frame that the training data was collected in. Since hardware and simulation are significantly different, this coordinate frame is not the same between sim and real. We build a transformation that converts hardware measured poses to the simulation frame of reference, which is then used to create the instance tokens. This ensures that there is no sim-to-real gap in object positions, reducing the challenges involved in applying such a simulation trained policy to hardware. In this section we describe how we convert the real world coordinates to simulation frame coordinates for running the trained TTP policy on a Franka arm.

We use the semantic work area in simulation and hardware to transform the hardware position coordinates to simulation position coordinates. We measure the extremes of the real workspace by manually moving the robot to record positions and orientations that define the extents of the workspace for the table. The extents of the drawers are measured by placing ARTag markers. We build 3 real-to-sim transformations using the extents for counter, top rack and bottom rack:
Let $X \in \mathbb{R}^{3\times N}$ contain homogeneous $xz-$ coordinates of a work area, along its column, as follows: 
\begin{equation}
    X = \begin{bmatrix}
     x^{(1)} & x^{(2)} & \cdots\\
     z^{(1)} & z^{(2)} & \cdots\\
     1 & 1 & \cdots\\
     \end{bmatrix} 
     =  \begin{bmatrix}
     \boldsymbol{x}^{(1)} & \boldsymbol{x}^{(2)} & \cdots
     \end{bmatrix}
\end{equation}

As the required transformation from real to simulation involves scaling and translation only, we have 4 unknowns, namely, $\boldsymbol{a} = [\alpha_x, \alpha_y, x_{trans}, z_{trans}]$. 
Here $\alpha_x, \alpha_z$ are scaling factors and $x_{trans}, z_{trans}$ are translation offset for $x$ and $z$ axis respectively. 
To solve $ X_{sim} = A X_{hw}$, we need to find the transformation matrix $A = \hat{\boldsymbol{a}} =  \begin{bmatrix}
     \alpha_x & 0 & x_{trans}\\
     0 & \alpha_z & z_{trans} \\
     0 & 0 & 1 \\
     \end{bmatrix} $.

\begin{align}
    X_{sim} &= \hat{\boldsymbol{a}} X_{hw} \\
     \text{Rewriting the } & \text{ system of linear equations,} \\
     \implies 
     \begin{bmatrix}
     x^{(1)}_{sim}\\
     z^{(1)}_{sim}\\
     x^{(2)}_{sim} \\
     z^{(2)}_{sim} \\
     \vdots \\
     \end{bmatrix} &= \begin{bmatrix}
     x^{(1)}_{hw} & 0 & 1 & 0 \\
     0 & z^{(1)}_{hw} & 0 & 1 \\
     x^{(2)}_{hw} & 0 & 1 & 0 \\
     0 & z^{(2)}_{hw} & 0 & 1 \\
     \vdots & \vdots & \vdots & \vdots \\
     \end{bmatrix}  \boldsymbol{a}^T \\
\end{align}
Let the above equation be expressed as $Y_{sim} = Z_{hw} a^T$ where $Y_{sim} \in \mathbb{R}^{2N\times 1}$, $Z_{hw} \in \mathbb{R}^{2N \times 4}$, and $a^T \in \mathbb{R}^{4 \times 1}$.
Assuming we have sufficient number of pairs of corresponding points in simulation and real world, we can solve for $\boldsymbol{a}$ by least squares $a = (Z_{hw}^T Z_{hw})^{-1} Z_{hw}^T Y_{sim}$.
The height $y_{sim}$ is chosen from a look-up table based on $y_{hw}$.
Once we compute the transformation $A$, we store it for later to process arbitrary coordinates from real to sim, as shown below.

\begin{minted}{python}
def get_simulation_coordinates(xyz_hw: List[float], A: np.array) -> List:
    xz_hw = [xyz_hw[0], xyz_hw[2]]
    X_hw = get_homogenous_coordinates(xz_hw)
    X_sim_homo = np.matmul(A, X_hw)
    y_sim = process_height(xyz_hw[1]) 
    X_sim = [X_sim_homo[0]/X_sim_homo[2], y_sim, X_sim_homo[1]/X_sim_homo[2]]
    return X_sim
\end{minted}


The objects used in simulation training are different from hardware objects, even though they belong to the same categories. For example, while both sim and real have a small plate, the sizes of these plates are different. We can estimate the size of the objects based on actual bounding box from the segmentation pipeline. However, it is significantly out-of-distribution from the training data, due to object mismatch. So, we map each detected object to the nearest matching object in simulation and use the simulation size as the input to the policy. This is non-ideal, as the placing might differ for sim versus real objects. In the future, we would like to train with rich variations of object bounding box size in simulation so that the policy can generalize to unseen object shapes in the real world.

\section{Simulation Setup}
\begin{figure}
    \centering
    \includegraphics[width=\textwidth]{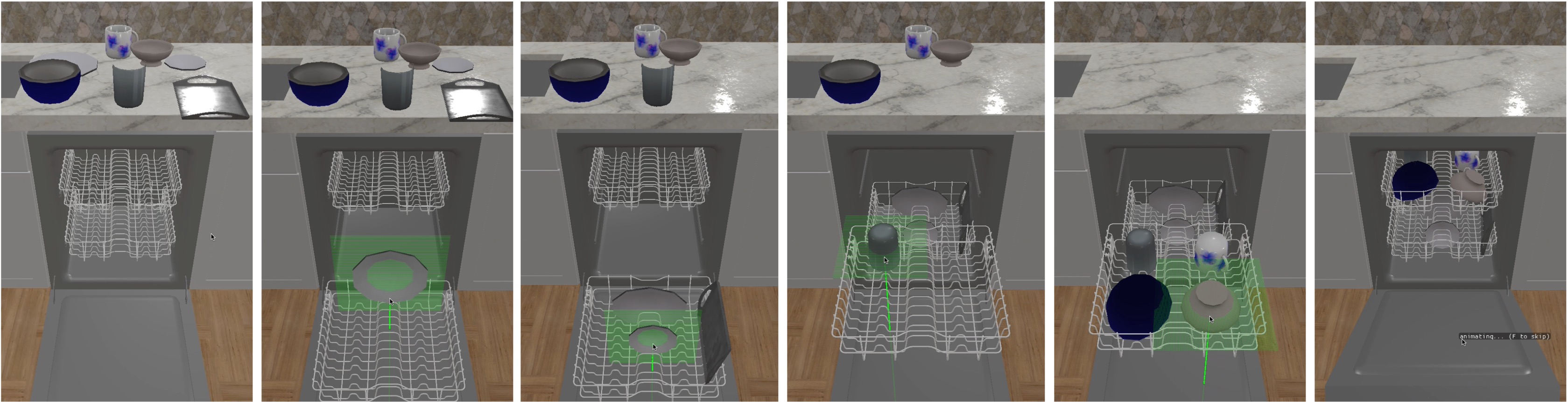}
    \caption{Human demonstration with point and click in simulation}
    \label{fig:human_demo_point_n_click}
\end{figure}
\paragraph{Dataset}
\label{appsubsec:dataset}
``Replica Synthetic Apartment 0 Kitchen" consists of a fully-interactive dishwasher with a door and two sliding racks, an adjacent counter with a sink, and a ``stage" with walls, floors, and ceiling.
We use selected objects from the ReplicaCAD \cite{szot2021habitat} dataset, including seven types of dishes (cups, glasses, trays, small bowls, big bowls, small plates, big plates) which are loaded into the dishwasher. Fig. \ref{fig:human_demo_point_n_click} shows a human demonstration recorded in simulation by pointing and clicking on desired object to pick and place.  
We initialize every scene with an empty dishwasher and random objects placed on the counter. Next, we generate dishwasher loading demonstrations, adhering to a given preference, using an expert-designed data generation script. 
Expert actions include, opening/closing dishwasher/racks, picking/placing objects in feasible locations or the sink if there are no feasible locations left. Experts differ in their preferences, and might choose different object arrangements in the dishwasher. 

\begin{wraptable}{r}{0.45\linewidth}
\centering
\vspace{-10pt}
\caption{Three example preferences for dishwasher loading. Rack order and their respective contents (ordered by preference).}
\footnotesize
\begin{tabular}{@{}l@{\hspace{1em}}l@{\hspace{1em}}l@{}}
\textit{First?} & \textit{Top}                                                                       & \textit{Bottom}                                                                           \\
\toprule
Top             & \begin{tabular}[c]{@{}l@{}}1. cups\\ 2. glasses\\ 3. small bowl\end{tabular}               & \begin{tabular}[c]{@{}l@{}}1. big plates\\ 2. small plates\\ 3. trays\\ 4. big bowls\end{tabular} \\
\midrule
Bottom          & \begin{tabular}[c]{@{}l@{}}1. cups\\ 2. glasses\\ 3. small bowl\\\end{tabular} & \begin{tabular}[c]{@{}l@{}}1. big plates\\ 2. small plates\\ 3. trays \\  4. big bowl \end{tabular}                \\
\midrule
Bottom          & \begin{tabular}[c]{@{}l@{}}1. small plate\\ 2. glasses\\ 3. cups\end{tabular}              & \begin{tabular}[c]{@{}l@{}}1. big bowls\\ 2. trays\\ 3. big plates\\ 4. small bowl\end{tabular}   \\ \bottomrule
\end{tabular}
\label{tab:examples_of_pref}
\end{wraptable}
\paragraph{Expert Preferences}
\label{subsec:pref_example}
 We define a preference in terms of expert demonstration `properties', like which rack is loaded first with what objects? There are combinatorially many preferences possible, depending on how many objects we use in the training set. For example, Table \ref{tab:examples_of_pref} describes the preferences of dishwasher loading in terms of three properties - first loaded tray, objects in top and bottom tray. Each preference specifies properties such as which rack to load first and their contents. In Table \ref{tab:examples_of_pref}, Preferences 1 \& 2 vary in the order of which rack is loaded first, while 2 \& 3 both load the bottom rack first with similar categories on top and bottom but with different orderings for these categories. Other preferences can have different combinations of objects loaded per rack.

To describe a preference, let there be $k$ properties, where each can take $m_k$ values respectively. For example, a property to describe preference can be which rack is loaded first, and this can take two values; either top or bottom rack. The total number of possible preferences is $G = \prod_{i=1}^{k} m_i.$ 

In our demonstration dataset, we have 100 unique sessions per preference. Each session can act as a prompt to indicate preference as well as provide situation for the policy. Each session is about $\sim 30$ steps long.  
With $7$ preferences, this leads to
$70,000 \times 30 = 2,100,000$ 
$\sim 2$ million total training samples, creating a relatively large training dataset from only 100 unique demonstrations per preference. Individual task preferences differ in the sequence of expert actions, but collectively, preferences share the underlying task semantics. 

\paragraph{Dynamically appearing objects}
To add additional complexity to our simulation environment, we simulate a setting with dynamically appearing objects later in the episode.
During each session, the scene is initialized with $p\%$ of maximum objects allowed. The policy/expert starts filling a dishwasher using these initialized objects. After all the initial objects are loaded and both racks are closed, new objects are initialized one-per-timestep to the policy. The goal is to simulate an environment where the policy does not have perfect knowledge of the scene, and needs to reactively reason about new information. 
The policy reasons on both object configurations in the racks, and the new object type to decide whether to `open a rack and place the utensil'  or  `drop the object in the sink'.

\section{Training}
\label{appsec:train}

In this Section we describe details of the different components of our learning pipeline.

\subsection{Baseline: GNN}
\paragraph{Architecture}
We use GNN with attention. The input consists of 12 dimensional attribute inputs (1D-timestep, 3D-category bounding box extents, 7D-pose, 1D-is object or not?) and 12 dimensional one-hot encoding for the preference. 

\begin{minted}{python}
input_dim: 24 
hidden_dim: 128
epochs: 200
batch_size: 32
\end{minted}

\paragraph{Optimizer}: Adam with $lr=0.01$ and weight\_decay$=1e-3$.

\paragraph{Reward function for GNN-RL}
\label{subsec:rewardRL}
Reward function for the RL policy is defined in terms of preference.
The policy gets a reward of +1 every time it predicts the instance to pick that has the category according to the preference order and whether it is placed on the preferred rack.

\subsection{Our proposed approach: TTP}
\paragraph{Architecture} 
We use a 2-layer 2-head Transformer network for encoder and decoder. The input dimension of instance embedding is 256 and the hidden layer dimension is 512. The attributes contribute to the instance embedding as follows: 
\begin{minted}{python}
C_embed: 16
category_embed_size: 64
pose_embed_size:  128
temporal_embed_size: 32 
marker_embed_size: 32
\end{minted}
For the slot attention layer at the head of Transformer encoder, we use: 
\begin{minted}{python}
num_slots: 50 
slot_iters: 3
\end{minted}

\paragraph{Optimizer}
We use a batch-size of 64 sequences. Within each batch, we use pad the inputs with 0 upto the max sequence length. 
Our optimizer of choice is SGD with momentum 0.9, weight decay 0.0001 and dampening 0.1. The initial learning rate is 0.01, with exponential decay of 0.9995 per 10 gradient updates.
We used early stopping with patience 100. 

\begin{figure}[t]
    \centering   
    \begin{subfigure}[b]{0.3\textwidth}
     \centering 
        \includegraphics[width=0.8\textwidth, trim={0 0 0 5cm},clip]{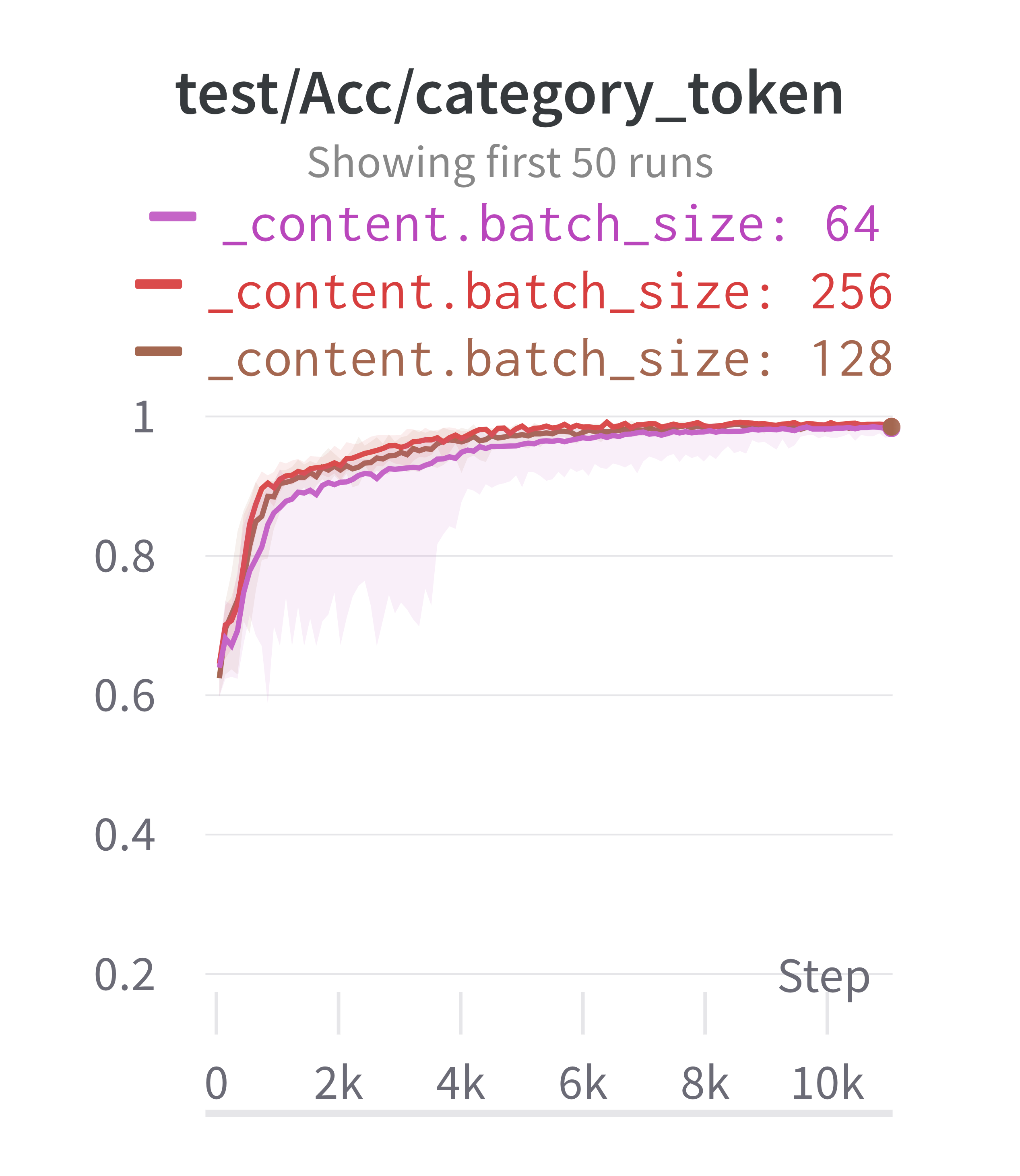}
        \caption{Category level accuracy grouped by batch size for prompt-situation training.}
        \label{fig:cataccbybs}
    \end{subfigure}
    \begin{subfigure}[b]{0.67\textwidth}
     \centering 
         \includegraphics[width=1.\textwidth, trim={3em, 0em, 0em, 4em}, clip]{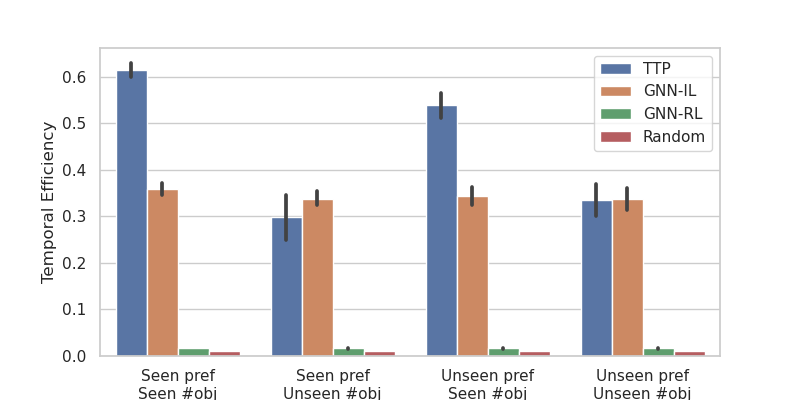}
         \caption{TE (SPL) metric for held-out test settings.}
         \label{fig:te}
    \end{subfigure}
\end{figure}
\subsection{Metrics}

In Section 3,
we presented packing efficiency (PE) and edit distance (ED) metrics collected on a policy rollout. We present additional metrics about training progress and rollout here.


\textbf{Category-token Accuracy} indicates how well the policy can mimic the expert's action, given the current state. We monitor training progress by matching the predicted instance to the target chosen in demonstration (Fig. \ref{fig:cataccbybs}). We see that TTP is able to predict the same category object to pick perfectly (accuracy close to $1.0$). However, this is a simpler setting that sequential decision making. During rollout, any error in a state could create a setting that is out-of-distribution for the policy. Thus, category token accuracy sets an upper bound for rollout performance, that is, while having high category token accuracy is necessary, it is not sufficient for high packing efficiency and inverse edit distance. 


 
 
\textbf{Temporal efficiency}: Just like SPL \cite{anderson2018evaluation} for navigation agents, we define the efficiency of temporal tasks in policy rollout, in order to study how efficient the agent was at achieving the task. 
For episode $i \in [1,..N]$, let the agent take $p_i$ number of high-level interactions to execute the task, and the demonstration consists of $l_i$ interactions for the initial state. We scale the packing efficiency $PE_i$ of the policy by the ratio of steps taken by expert versus policy.
Temporal efficiency is defined between 0 to 1, and higher is better. This value will be equal to or lower than the packing efficiency. This especially penalizes policies that present a `looping' behavior, such as repeatedly open/close dishwasher trays, over policies that reach a low PE in shorter episodes (for example, by placing most objects in the sink). Fig \ref{fig:te} shows the temporal efficiency or SPL over our 4 main held-out test settings. 


\section{Additional Ablation Experiments}
In Section 3.3, we presented ablation experiments over number of demonstrations per preference used for training, and the number of unique preferences used. In this Section, we present additional ablation experiments over the design of instance encodings in TTP. Additionally, we also present results where we increase the temporal context of TTP and study its effect on performance.

\subsection{Design of Instance Encoding}
\label{appsubsec:design_instance}

\begin{figure}[t]
\centering
\begin{minipage}{0.32\textwidth}
    \centering
    \includegraphics[width=1\textwidth,trim={0.5em 1.5em 0.5em 3.5em},clip]{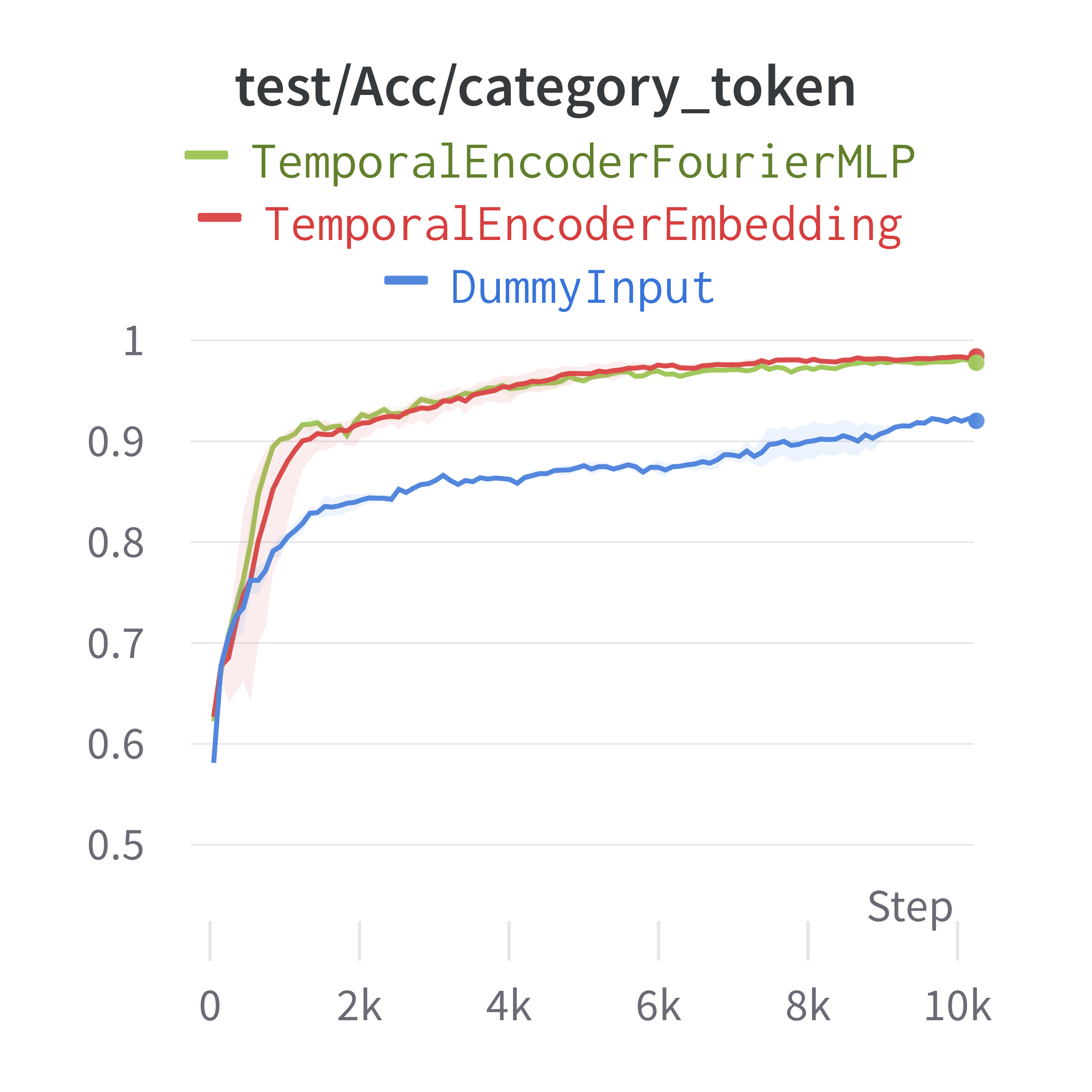}
    \subcaption{Temporal encoding}
    \label{fig:ablation_temporal}
\end{minipage}
\begin{minipage}{0.32\textwidth}
    \centering
    \includegraphics[width=1\textwidth,trim={0cm 0.5cm 0cm 1.5cm},clip]{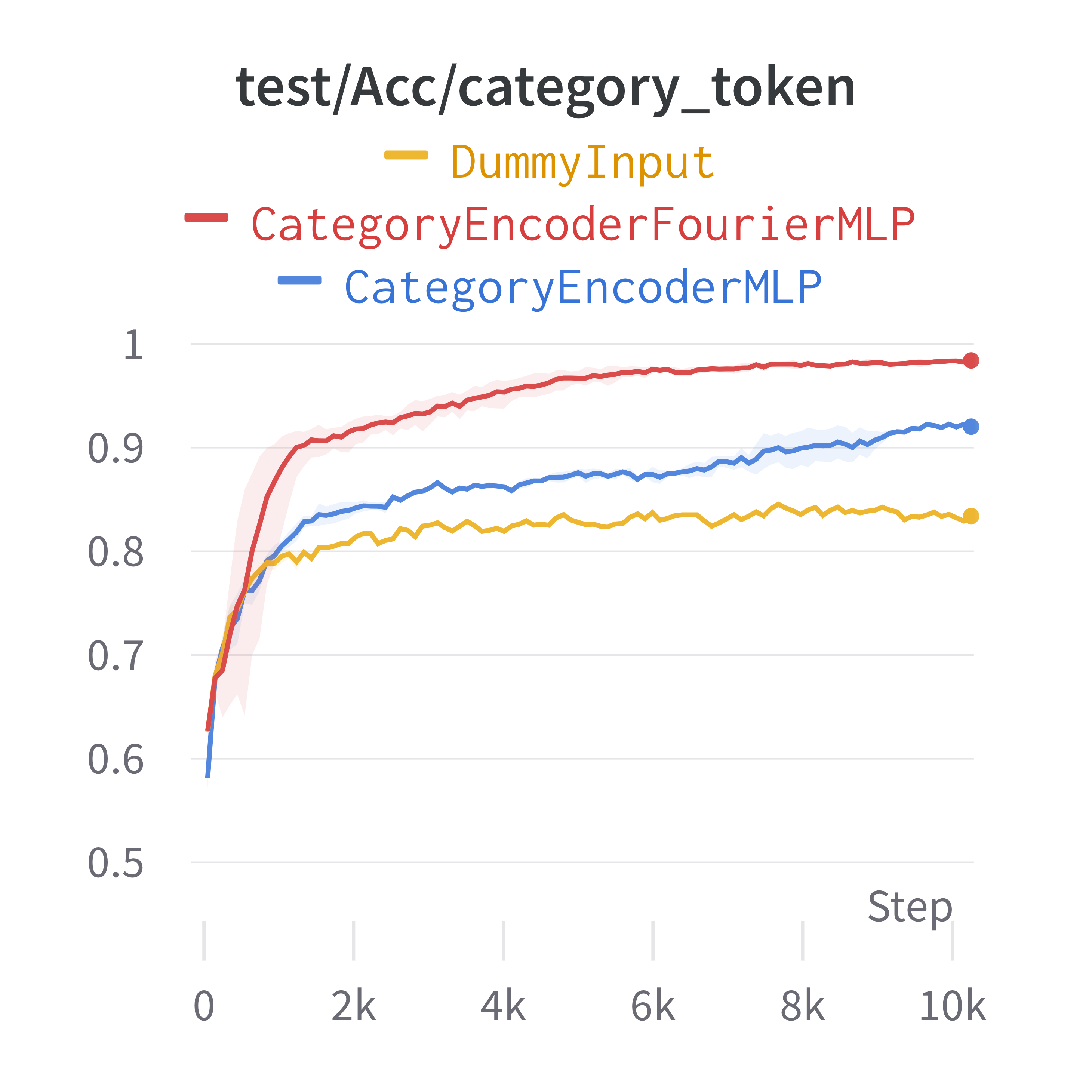}
    \subcaption{Category encoding}
    \label{fig:ablation_category}
\end{minipage}
\begin{minipage}{0.32\textwidth}
    \centering
    \includegraphics[width=1\textwidth,trim={0cm 0.5cm 0cm 1.5cm},clip]{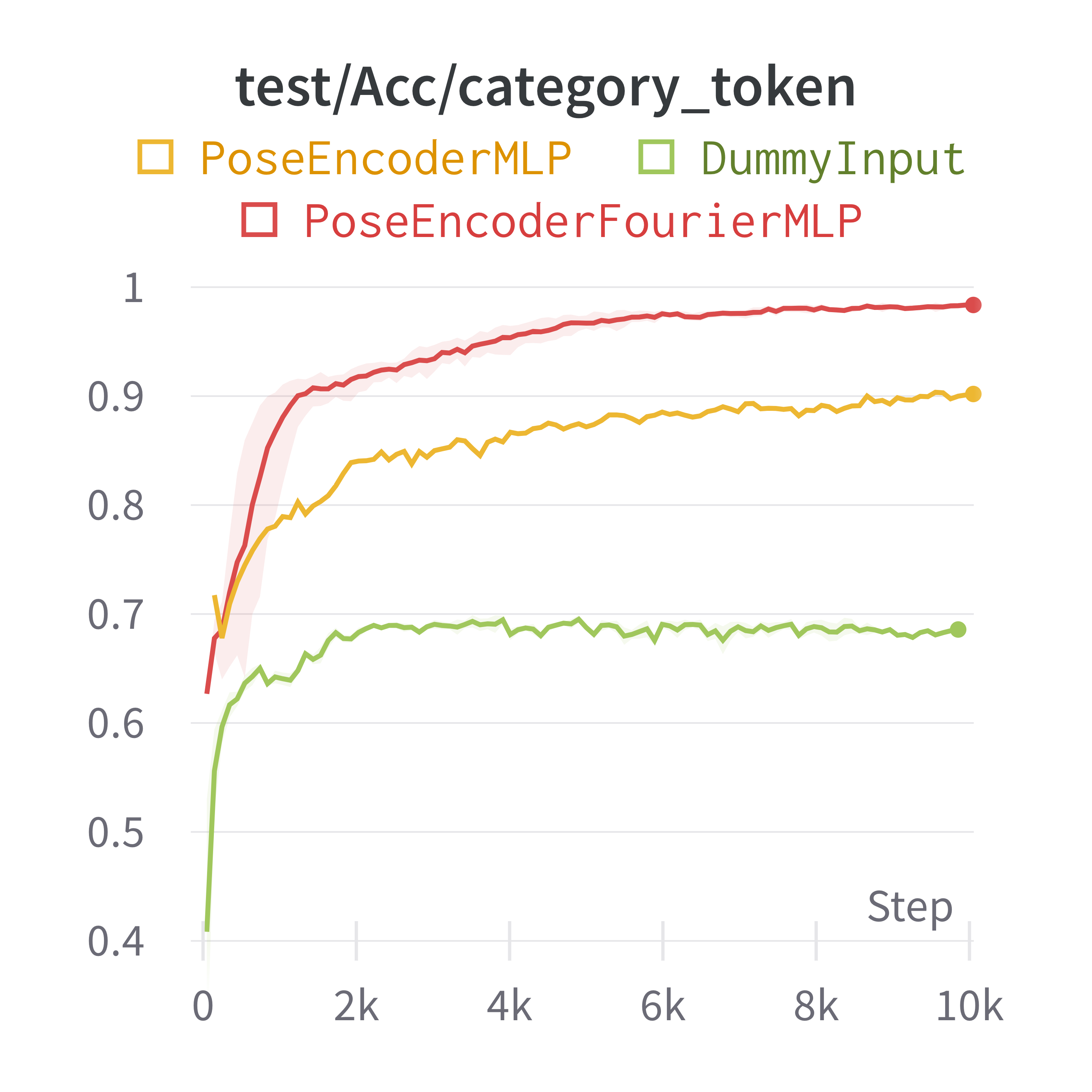}
    \subcaption{Pose encoding}
    \label{fig:ablation_pose}
\end{minipage}
\caption{[Left-to-Right] Comparing different design choices of attribute encoders in terms of category token accuracy on held-out test prompt-situation session pairs. 
}
\label{fig:ablations}
\end{figure}





\paragraph{How much does \textbf{temporal} encoding design matter?}
Fig. \ref{fig:ablation_temporal} shows that learning an embedding per timestep or expanding it as fourier transformed vector of sufficient size achieves high success. On the other hand, having no timestep input shows slightly lower performance. Timestep helps in encoding the order of the prompt states. The notion of timestep is also incorportated by autoregressive masking in both the encoder and the decoder.
\vspace{-0.5em}



\paragraph{How much does \textbf{category} encoding design matter?}
In our work, we represent category as the extents of an objects' bounding box. 
An alternative would be to denote the category as a discrete set of categorical labels. 
Intuitively, bounding box extents captures shape similarity between objects and their placement implicitly, which discrete category labels do not. Fig. \ref{fig:ablation_category} shows that fourier transform of the bounding box achieves better performance than discrete labels, which exceeds the performance with no category input.
\vspace{-0.5em}
\paragraph{How much does \textbf{pose} encoding design matter?}
We encode pose as a 7-dim vector that includes 3d position and 4d quaternion.
Fig. \ref{fig:ablation_pose} shows that the fourier transform of the pose encoding performs better than feeding the 7 dim through MLP. 
Fourier transform of the pose performs better because such a vector encodes the fine and coarse nuances appropriately, which otherwise either require careful scaling or can be lost during SGD training.  





\subsection{Markov assumption on the current state in partial visibility scenarios} 

\label{appsubsec:context_history}
\begin{wrapfigure}{r}{0.505\textwidth}
\vspace{-0.5em}
    \centering
    \includegraphics[width=0.45\textwidth]{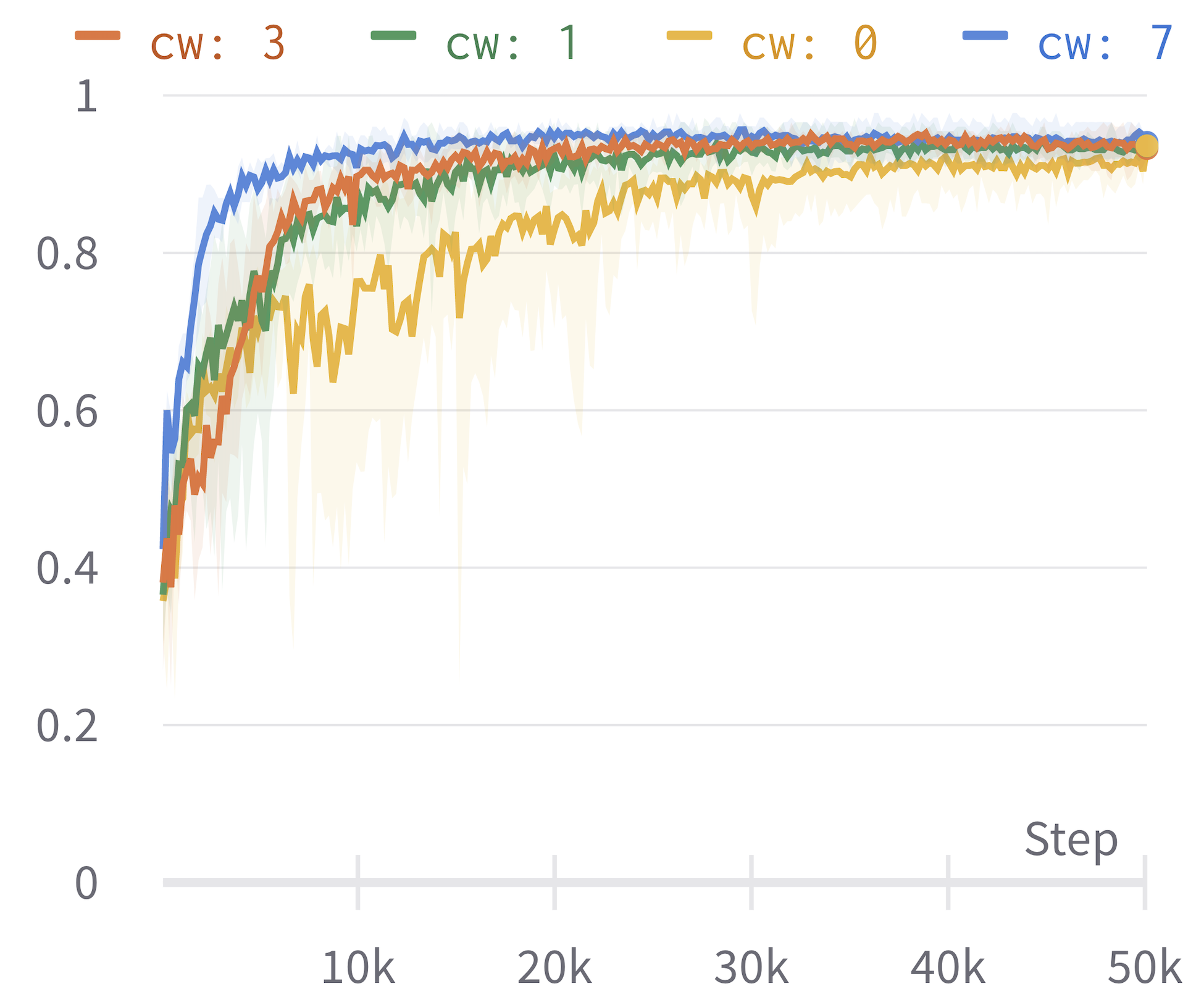}
    \caption{Category level accuracy for single preference training with varying context windows. }
    \label{fig:context_history}
    \vspace{-1em}
\end{wrapfigure}
\begin{figure}[t]
    \centering
    \includegraphics[width=0.7\textwidth]{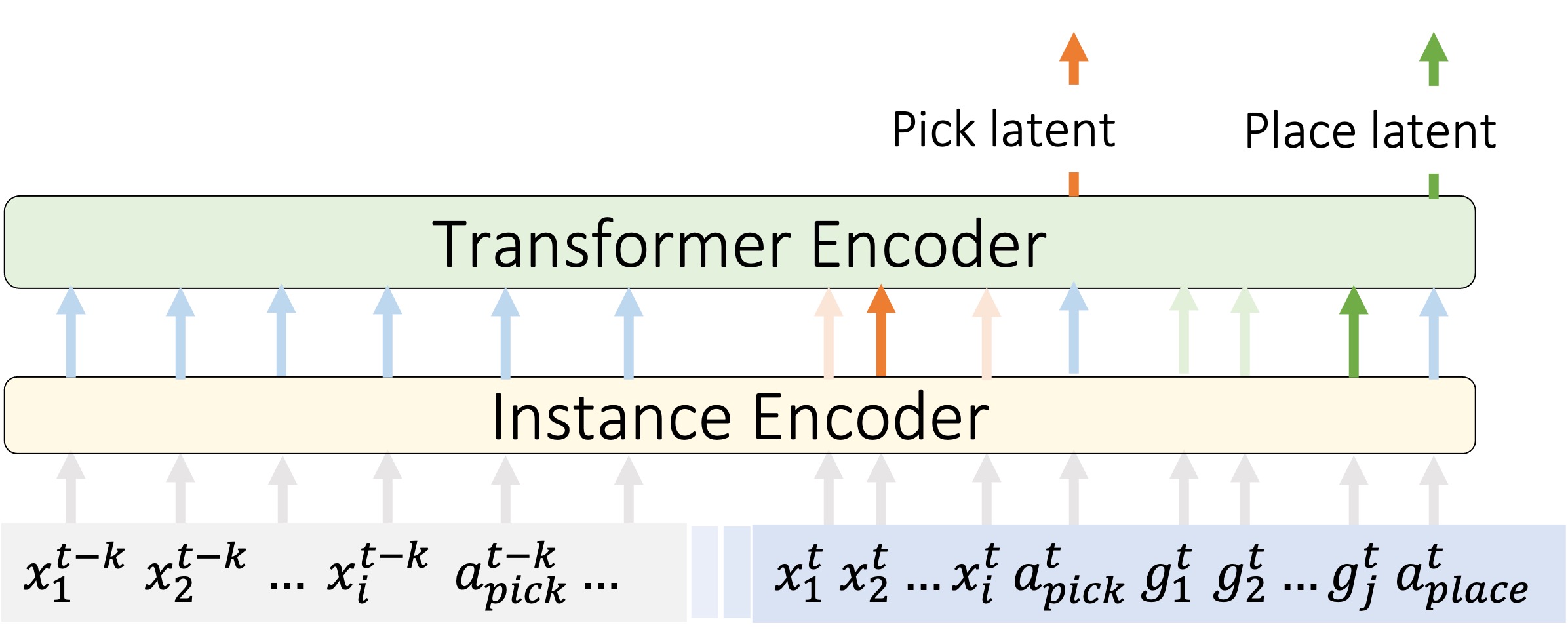}
    \caption{Processing with previous Context History $k$}
    \label{fig:explain_cw}
\end{figure}

Dynamic settings, as used in our simulation, can be partially observable. For example, when the rack is closed, the policy doesn't know whether it is full or not from just the current state. If a new object arrives, the policy needs to decide between opening the rack if there is space, or dropping the object in sink if the rack is full. In such partially observed settings, the current state may or may not contain all the information needed to reason about the next action. However, given information from states in previous timesteps, the policy can decide what action to take (whether to open the rack or directly place the object in the sink). 
To this end, we train a single preference pick only policy for different context history. 
As shown in Fig. \ref{fig:explain_cw}, context window of size $k$ processes the current state as well as $k$ predecessor states, that is, in total $k+1$ states. While larger context window size learns faster, the asymptotic performance for all context windows converges in our setting.

Let context history $k$ refer to the number of previous states included in the input.  
Then the input is a sequence of previous $k$ states' instances (including the current state), as shown in Fig. \ref{fig:explain_cw}.
Fig \ref{fig:context_history} shows that TTP gets $> 90\%$ category level prediction accuracy in validation for all context windows. While larger context windows result in faster learning at the start of the training, the asymptotic performance of all contexts is the same. This points to the dataset being largely visible, and a single context window capturing the required information. In the future, we would like to experiment with more complex settings like mobile robots, which might require a longer context.

\section{Limitations and Future scope}
In Section 6, we briefly discussed the limitations and risks. Here we enlist more details and highlight future directions. 
\vspace{-0.5em}
\paragraph{Pick grasping depends on accurate segmentation and edge detection} Grasping policy depends on quality of segmentation and edge detection of the selected object. Due to noise in calibration, shadows and reflections, there are errors in detecting the correct edge to successfully grasp the object. 
For example, it is hard to grasp a plate in real setting. Plate is very close to the ground and the depth cameras cannot detect a clean edge for grasping. Therefore, in our work, we place the plate on an elevated stand for easy grasping. Grasping success also depends on the size and kind of gripper used.
\vspace{-0.5em}
\paragraph{Placement in real setting} For placement, the orientation of final pose is often different from initial pose and  may require re-grasping. The placement pose at final settlement is different from the robot's end-effector pose while releasing the object from its grasp. Similar to picking, placement accuracy will largely depend on approperiate size and shape of gripper used. Due to these reasons, placement in real world is an open challenging problem and we hope to address this future work.  
\vspace{-0.5em}
\paragraph{Hardware pipeline issues due to calibration} 
The resulting point cloud is generated noisy due to two reasons. First, incorrect depth estimation due to camera hardware, lighting conditions, shadows and reflections. Second, any small movements among cameras that affects calibration. 
If we have a noisy point cloud, it is more likely to have errors in subsequent segmentation and edge detection for grasp policy.
Having sufficient coverage of the workspace with cameras is important to mitigate issues due to occlusions and incomplete point clouds. 
\vspace{-0.5em}
\paragraph{Incomplete information in prompt} The prompt session may not contain all the information to execute on the situation. For example, in a prompt session there might be no large plates seen, which in incomplete or ambiguous information for the policy. This can be mitigated by ensuring complete information in prompt demo or having multiple prompts in slightly different initialization. 

\end{document}